\documentclass[10pt,twocolumn,letterpaper]{article}
\usepackage[pagenumbers]{cvpr}
\usepackage{bm}
\usepackage{amsthm}
\usepackage{microtype}
\usepackage{float}

\newtheorem{proposition}{Proposition}

\floatstyle{ruled}
\newfloat{algorithm}{tbp}{loa}
\floatname{algorithm}{Algorithm}
\floatstyle{plain}

\makeatletter
\renewcommand{\paragraph}{%
  \@startsection{paragraph}{4}{\z@}%
    {2.0ex \@plus 1ex \@minus .2ex}%
    {-1em}%
    {\normalfont\normalsize\bfseries}}
\makeatother
\definecolor{cvprblue}{rgb}{0.21,0.49,0.74}
\usepackage{hyperref}

\title{FedQCal: Quadratic Programming with\texorpdfstring{\\}{ }%
Dual-Space Calibration for One-Shot Federated Learning}
\newcommand{\affmark}[1]{\texorpdfstring{\textsuperscript{#1}}{}}
\newcommand{\paperauthors}{Shunsei Koshibuchi\affmark{1}, Hiroshi Kera\affmark{1,2}, and Kazuhiko Kawamoto\affmark{1}}
\author{\paperauthors\\[3pt]
  {\normalsize \textsuperscript{1}Chiba University \qquad
               \textsuperscript{2}National Institute of Informatics}}
\hypersetup{
  breaklinks=true,
  colorlinks=true,
  allcolors=cvprblue,
  pdftitle={\thetitle},
  pdfauthor={\paperauthors}
}

\begin{document}
\raggedbottom
\maketitle
\begin{abstract}
One-shot federated learning must combine heterogeneous client models after a single upload.
Different local class distributions make it difficult
to decide which client updates to retain.
We propose FedQCal, which uses a reference model trained on a small labeled server set
to define alignment requirements for aggregation.
The key idea is to retain sufficient reference alignment
while avoiding unnecessarily large aggregation coefficients.
A minimum-norm quadratic program (QP) enforces these requirements.
Classification loss on the same server examples selects a merge through two dual variables,
so the alignment thresholds need not be prescribed in advance.
We prove that the selected coefficients remain exact QP solutions at their induced thresholds.
A perturbation bound explains how coefficient norm limits the amplification of projection errors.
Experiments on CIFAR-10, CIFAR-100, and SVHN show gains over
one-round FedAvg in most settings.
Ablations support reference-guided weighting and coefficient calibration,
even with one labeled server image per class.
\end{abstract}

\section{Introduction}
\label{sec:intro}
Federated learning (FL) combines knowledge from clients without collecting their training
data~\cite{mcmahan2017communication,kairouz2021advances}.
One-shot federated learning limits communication to a single upload of local
models~\cite{guha2019one}.
Clients adapt a shared pre-trained model, and the server merges their updates into one
classifier.
The difficulty is that clients may observe very different class distributions.
An update useful for one client may be less suitable for the full task, and no further round
of client training is available to correct the merge.

Existing approaches use different criteria to combine client updates.
Sample-weighted averaging reflects local dataset sizes~\cite{mcmahan2017communication},
while TIES-Merging uses agreement among parameter updates~\cite{yadav2024ties}.
FedAWA adapts aggregation weights using alignment between client updates
and their aggregate, without auxiliary data~\cite{shi2025fedawa}.
Data volume and update agreement, however, do not directly measure
whether an update benefits the full classification task.
Server supervision provides task information:
FedLAW learns relative client weights and an overall aggregation scale~\cite{li2023fedlaw},
while Feddle optimizes combinations of client updates using server data
during federated training~\cite{zhu2026guided}.
When only a few labeled server examples are available,
assessments of which updates are useful can be uncertain.
Large coefficients can make the merge sensitive to errors in these assessments,
while uniformly shrinking coefficients also suppresses useful adaptation.

\begin{figure*}[t]
\centering
\includegraphics[width=0.92\textwidth]{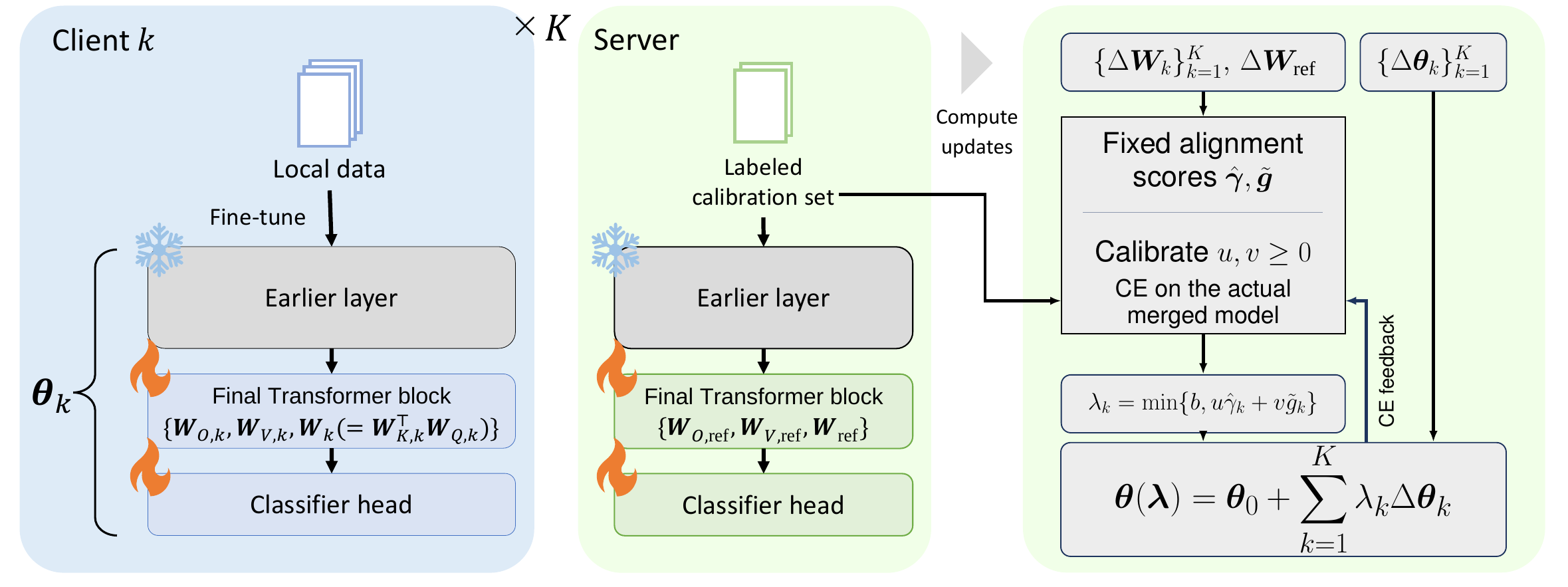}
\caption{\textbf{FedQCal: a reference defines the merge family;
classification loss selects a member.}
After a single client upload, the server fits a reference and computes
two fixed alignment scores per client.
The dual coordinates $u,v$ map these scores to minimum-norm QP solutions.
Cross-entropy (CE) on the merged network updates the coordinates using
the same calibration examples; the reference and client updates remain fixed.
Snowflakes and flames indicate frozen and trainable components during
client and reference training.
For clarity, only attention-related quantities are shown inside the
final Transformer blocks; training updates the entire final block
and classifier.}
\label{fig:overview}
\end{figure*}

We address this trade-off with \emph{FedQCal}.
The server adapts the shared pre-trained model to its
labeled examples to obtain a reference model.
The reference adaptation gives a common basis for assessing client updates learned from
different class distributions.
FedQCal expresses agreement with this reference as constraints
and uses quadratic programming (QP) to find the smallest coefficient norm that satisfies them.
This couples the two goals: the constraints specify what to preserve,
and the objective controls how strongly the merge depends on the client updates.
FedQCal uses classification loss on the same server examples to
adjust the required degree of agreement.
Figure~\ref{fig:overview} summarizes the procedure.
We establish the minimum-norm QP structure of the calibrated coefficients
and bound the effect of reference and summary errors on aggregate projection.
Ablations show that reference-guided weighting outperforms a random-direction control
in mean accuracy and that calibration improves the initial merge,
even with one labeled server example per class.

Our contributions are as follows:
\begin{itemize}
\item We propose FedQCal, a reference-guided method for one-shot federated aggregation.
Its dual formulation uses classification loss to select a merge
within a family of minimum-norm QP solutions.
\item We prove that the calibrated coefficients solve a minimum-norm QP at the selected
alignment requirements.
A projection perturbation bound explains how coefficient-norm control limits the effect of
reference and summary errors.
\item \begin{samepage}Experiments on three datasets with 10 and 100 clients show gains over
FedAvg in most settings.
Ablations support reference-guided weighting and coefficient calibration
with limited server supervision.\par
\end{samepage}
\end{itemize}

\section{Related Work}
\label{sec:related}
\paragraph{Federated aggregation.}
One-shot FL restricts communication to a single upload round~\cite{guha2019one},
making the server's use of the uploaded information central to the final model.
One-round FedAvg averages local parameters according to sample
counts~\cite{mcmahan2017communication}, while ensembling retains local
predictors~\cite{jeon2024revisiting}.
Distillation combines client knowledge through predictions or synthetic examples:
DOSFL transmits distilled data~\cite{zhou2020distilled}; FedDF and FedFTG train a server model from client predictions
in iterative FL~\cite{lin2020ensemble,zhang2022fine};
and DENSE generates examples for one-shot distillation~\cite{zhang2022dense}.
FAFI instead improves compatibility before aggregation through local feature alignment
and prototype learning~\cite{zeng2025fafi}.
These approaches address the information lost when heterogeneous predictors are combined,
using different client training and server reconstruction procedures.

A second group transmits data-derived statistics to support aggregation.
AFL and SAFLe construct analytic predictors from pretrained
representations~\cite{gao2024afl,bacellar2026safle}.
FedLPA uses layer-wise posteriors~\cite{wang2024fedlpa},
FedFisher uses local Fisher matrices~\cite{jhunjhunwala2024fedfisher},
and MA-Echo uses data-dependent projections~\cite{su2023oneshot}.
FedPFT transfers Gaussian mixture parameters~\cite{beitollahi2024parametric},
while GH-OFL uses class counts and feature moments~\cite{turazza2026gaussian}.
FedQCal studies the setting in which the server receives model updates
and holds a small labeled calibration set.
It derives the alignment scores from those models,
without requesting client-side feature moments, curvature estimates, or synthetic data.
The server can then evaluate candidate merges locally within the single-upload protocol.

\paragraph{Model merging.}
Shared initialization makes parameter-space merging possible:
model soups average fine-tuned weights~\cite{wortsman2022soups},
and task arithmetic expresses adaptation as parameter differences,
called task vectors~\cite{ilharco2023editing,tao2025task}.
The aggregation criterion determines which aspects of the updates are preserved.
TIES-Merging uses trimming and sign agreement~\cite{yadav2024ties};
Fisher merging weights parameters using posterior precision~\cite{matena2022fisher};
and RegMean uses activation statistics in layer-wise regression~\cite{jin2023dataless}.
These criteria provide structure for merging models beyond uniform averaging.

Alignment also plays a role in theoretical studies of task-vector operations.
Li et al.~\cite{li2025when} derive generalization guarantees for structured binary tasks in
a single-head, one-layer Transformer; their out-of-domain analysis assumes orthonormal
source discriminative patterns.
This work motivates relating task alignment to coefficient selection,
but its assumptions do not directly describe multiclass Vision Transformer (ViT) calibration.
We therefore analyze the implemented coefficient map itself:
its exact correspondence to QP solutions, its reference projection,
and its sensitivity to summary and reference errors.
The experiments assess whether selecting among these QP solutions improves classification.

\paragraph{Learned aggregation weights.}
Aggregation weights can also be learned from model updates or server data.
FedLAW learns relative aggregation weights and a global scaling factor from a server proxy
set during federated training~\cite{li2023fedlaw}.
FedAWA uses client-vector directions to adapt weights in iterative FL without auxiliary data~\cite{shi2025fedawa}.
AdaMerging minimizes prediction entropy on unlabeled target samples to learn task-wise or
layer-wise coefficients~\cite{yang2024adamerge}.
Feddle uses server data to search combinations of client updates,
with fallback regularization toward baseline coefficients in its out-of-domain setting,
and returns the refined model for further federated training~\cite{zhu2026guided}.
Our contribution concerns the structure of the coefficient search:
two reference-alignment scores define a family of minimum-norm QP solutions,
and calibration cross-entropy selects within this family through its dual coordinates.
The reference supplies the geometry of the candidate merges,
while predictions supply the criterion for choosing among them.

\section{Reference-Guided QP Calibration}
\label{sec:method}
Client updates reflect different local class distributions,
so aggregation needs a common criterion for which changes to retain.
FedQCal uses a server-trained reference to define alignment requirements
and a QP to meet them with minimum coefficient norm.
This controls sensitivity of aggregate projection to errors in the alignment criterion.
Classification loss then selects how much alignment to require,
while preserving the QP's minimum-norm property (Figure~\ref{fig:overview}).
Algorithm~\ref{alg:fedqp} gives the server-side procedure.

\subsection{One-shot protocol and merged model}
There are $K$ clients with local datasets $\mathcal D_k$ of size $N_k$.
Each client adapts shared parameters $\bm{\theta}_0$,
including a common task-specific classifier initialization, to local parameters $\bm{\theta}_k$
and uploads $\Delta\bm{\theta}_k=\bm{\theta}_k-\bm{\theta}_0$ once.
We adapt the last transformer block and classifier; preceding layers remain frozen.
The server constructs a single model as
\begin{equation}
 \bm{\theta}(\bm{\lambda})=\bm{\theta}_0+\sum_{k=1}^K\lambda_k\Delta\bm{\theta}_k,
 \quad \bm{\lambda}\in[-b,b]^K,
 \label{eq:merge}
\end{equation}
where $\bm{\lambda}=(\lambda_1,\ldots,\lambda_K)^\top$ contains the client coefficients,
$b>0$ bounds their magnitudes, and $\top$ denotes transposition.
Each coefficient weights a client's complete trainable update, including the classifier.
The coefficients need not sum to one, so they determine both the relative client weights
and the overall adaptation scale.

\begin{algorithm}[t]
\caption{FedQCal server calibration}
\label{alg:fedqp}
\small
\textbf{Inputs:} shared model $\bm{\theta}_0$, fixed client updates and counts
$\{\Delta\bm{\theta}_k,N_k\}_{k=1}^K$, labeled set $\mathcal D_{\mathrm{cal}}$,
cap $b>0$,
iteration budget $T$, minimum step size $\eta_{\text{min}}>0$ and Armijo backtracking settings.\par
\textbf{Output:} one merged model.
\begin{enumerate}\setlength{\itemsep}{1pt}\setlength{\parskip}{0pt}
\item Fine-tune $\bm{\theta}_0$ on $\mathcal D_{\mathrm{cal}}$ to obtain the reference $\bm{\theta}_{\mathrm{ref}}$.
\item Compute the reference summary $\Delta\bm W_{\mathrm{ref}}$.
If it is zero, return the sample-weighted FedAvg model.
\item Compute client summaries and the fixed score matrix $\bm C$ using
Eq.~\eqref{eq:rho}--\eqref{eq:scores}.
If $\bm C=\bm0$, return the sample-weighted FedAvg model.
\item Initialize $\bm a_0$ using Eq.~\eqref{eq:initialization} and set $t=0$.
\item While $t<T$, form
$\bm{\lambda}_t=\operatorname{clip}_{[-b,b]}(\bm C^\top\bm a_t)$
and evaluate calibration CE on $\bm{\theta}(\bm{\lambda}_t)$.
Backpropagate through $\bm{\lambda}_t$ to the two coordinates $\bm a_t$.
Form the direction
\[
\bm d_t=
\Pi_{\mathbb R_+^2}
\left(\bm a_t-\nabla_{\bm a_t}L_{\mathrm{cal}}(\bm{\lambda}_t)\right)-\bm a_t.
\]
Apply Armijo backtracking along $\bm d_t$,
starting from $\eta=1$ and repeatedly halving $\eta$
until a finite, strictly decreasing trial satisfies the Armijo condition
or $\eta<\eta_{\text{min}}$.
If $\eta<\eta_{\text{min}}$, stop calibration and retain $\bm a_t$.
Otherwise, accept
\[
\bm a_{t+1}=\bm a_t+\eta\bm d_t
\]
and increment $t$.
Here $\nabla_{\bm a_t}$ denotes the gradient with respect to $\bm a_t$,
and $\Pi_{\mathbb R_+^2}$ replaces negative coordinates by zero.
\item Return $\bm{\theta}_0+\sum_k[\bm{\lambda}(\bm a_t)]_k\Delta\bm{\theta}_k$.
\end{enumerate}
Only the two coordinates change during calibration; the reference
and client updates stay fixed.
\end{algorithm}

The server holds a small labeled calibration set $\mathcal D_{\mathrm{cal}}$,
disjoint from client training and test data.
It first fine-tunes $\bm{\theta}_0$ on this set to obtain a reference model $\bm{\theta}_{\mathrm{ref}}$.
The reference update provides a common direction for scoring client updates.
The server selects the required alignment using the merged classifier's
cross-entropy (CE) on the same examples.
The reference, client updates, and scores remain fixed during coefficient selection.
Only the original client updates enter Eq.~\eqref{eq:merge}.

\begin{figure*}[t]
\centering
\includegraphics[width=\textwidth]{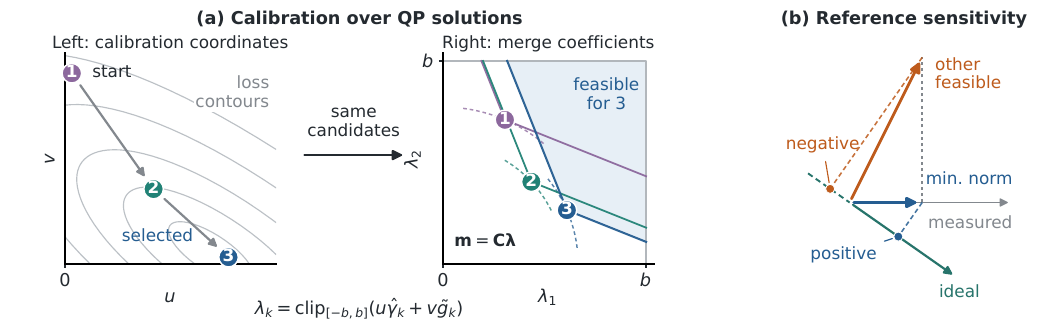}
\caption{\textbf{Calibration preserves the QP structure.}
(a) Left: a toy classification loss decreases from candidate 1 to 3;
the arrows illustrate a candidate sequence.
Right: each candidate minimizes the coefficient norm for its own requirements
$\bm m=\bm C\bm{\lambda}$.
Colors identify constraint boundaries; dashed arcs indicate equal norms.
Only candidate 3's feasible set is shaded.
(b) Equal aggregate alignment scores can hide different sensitivities to reference error.
Here, weighted summaries project positively (minimum-norm coefficients, blue)
or negatively (other feasible coefficients, orange) onto the ideal direction.}
\label{fig:geometry}
\end{figure*}

\subsection{Measuring alignment with the reference}
\label{sec:scores}
We compare how client and reference training change the key--query interaction
in the last attention block.
The product of these projections provides an input-independent matrix
that describes the bilinear interaction underlying attention scores:
\begin{equation}
 \bm W(\bm{\theta})=\bm W_K(\bm{\theta})^\top\bm W_Q(\bm{\theta}),
 \quad \bm W_0=\bm W(\bm{\theta}_0),
\end{equation}
where $\bm W_K$ and $\bm W_Q$ are the key and query projection matrices.
The same key/query and head convention is used for every model.

The server computes two alignment scores for each client:
normalized agreement $\hat\gamma_k$ and projection magnitude $\tilde g_k$.
Normalized agreement measures directional similarity independently of matrix magnitude.
Projection magnitude measures the amount of change along the reference direction.

\begin{samepage}
Let $\Delta\bm W_k=\bm W(\bm{\theta}_k)-\bm W_0$
and $\Delta\bm W_{\mathrm{ref}}=\bm W(\bm{\theta}_{\mathrm{ref}})-\bm W_0$.
These matrices summarize parameter changes, rather than attention probabilities.
For nonzero summaries, let $\rho_k$ be their cosine similarity
with negative values set to zero:
\begin{align}
 \rho_k&=\max\left\{0,
 \frac{\left\langle \Delta\bm W_k,\Delta\bm W_{\mathrm{ref}}\right\rangle_F}
 {\|\Delta\bm W_k\|_F\|\Delta\bm W_{\mathrm{ref}}\|_F}\right\},\label{eq:rho}\\
 \hat\gamma_k&=\sqrt{\rho_k},\quad
 \tilde g_k=\|\Delta\bm W_k\|_F\rho_k,\label{eq:scores}
\end{align}
where $\left\langle \bm X,\bm Y\right\rangle_F=\sum_{i,j}X_{ij}Y_{ij}$ is the Frobenius inner product
and $\|\bm X\|_F=\sqrt{\left\langle \bm X,\bm X\right\rangle_F}$ is the Frobenius norm.
\end{samepage}
The magnitude score is the positive part of the projection along the reference direction.
Nonpositive reference projections receive both scores zero,
restricting the coefficient family to clients that support the reference direction.

\subsection{From alignment requirements to a QP solution family}
\label{sec:qp}
The scores quantify support for the reference direction.
Large coefficients can amplify the effect of reference and summary errors
on the aggregate projection.
FedQCal therefore seeks the smallest coefficient norm that attains a required alignment.
Let $\hat{\bm\gamma}=(\hat\gamma_k)_{k=1}^K$ and $\tilde{\bm g}=(\tilde g_k)_{k=1}^K$,
and collect them as the rows of $\bm C=[\hat{\bm\gamma}^\top;\tilde{\bm g}^\top]\in\mathbb{R}^{2\times K}$.
For minimum aggregate scores $\bm m=(m_1,m_2)^\top$, the QP is
\begin{equation}
 \min_{\bm{\lambda}\in[-b,b]^K}\ \tfrac12\|\bm{\lambda}\|_2^2
 \quad\text{s.t.}\quad \bm C\bm{\lambda}\ge\bm m,
 \label{eq:qp}
\end{equation}
where $\|\bm x\|_2=\sqrt{\sum_i x_i^2}$ is the Euclidean norm
and vector inequalities are interpreted componentwise.
The constraints specify the required normalized agreement and projection magnitude,
keeping the norm objective from discarding the adaptation needed to meet them.
Alignment requirements alone can admit many coefficient vectors.
The objective resolves this ambiguity by selecting a unique minimum-norm solution,
the feasible point closest to the origin,
as illustrated on the right of Figure~\ref{fig:geometry}(a).
Conversely, minimizing the norm without alignment requirements would return zero coefficients
and leave the shared initialization unchanged.

The QP determines the merge once $\bm m$ is specified.
The scores measure agreement between attention updates,
but do not tell us which requirements produce accurate predictions.
The labeled server examples provide a direct criterion:
we evaluate the classification loss of the complete merged model.
Our aim is therefore to select a low-loss model among QP solutions,
while retaining the minimum-norm property.

The two alignment constraints provide a way to express these candidates explicitly.
Let $\bm a=(u,v)^\top$ collect their nonnegative Lagrange multipliers,
with $u$ for normalized agreement and $v$ for projection magnitude.
For fixed multipliers, the Lagrangian separates into one scalar quadratic per client.
Without the coefficient bound, each quadratic is minimized
when the coefficient equals the weighted sum of that client's two scores.
Enforcing the bound clips this value:
\begin{equation}
 \lambda_k(\bm a)=\operatorname{clip}_{[-b,b]}(u\hat\gamma_k+v\tilde g_k),
 \quad u,v\ge0,
 \label{eq:coefficient_map}
\end{equation}
Here $\lambda_k(\bm a)$ is the $k$th entry of the coefficient vector $\bm{\lambda}(\bm a)$.
The clipping operation is $z\mapsto\min\{b,\max\{-b,z\}\}$
and acts componentwise on vectors.
While clipping remains inactive, multiplying both coordinates
by the same positive factor changes only the overall adaptation scale.
Changing their ratio can also change the relative client weights.

Figure~\ref{fig:geometry}(a) illustrates this two-coordinate representation
of the QP solution family.
Each choice of $(u,v)$ specifies merge coefficients and the corresponding
alignment requirements $\bm m(\bm a)=\bm C\bm{\lambda}(\bm a)$.
We select among these candidates using classification loss,
without a threshold grid or repeated QP solves.

\subsection{Selecting a merge through classification loss}
\label{sec:dualce}
We now use the parameterization in Eq.~\eqref{eq:coefficient_map} to select a merged model
through its predictions on the calibration examples.
The calibration cross-entropy is
\begin{equation}
 L_{\mathrm{cal}}(\bm{\lambda})=\frac1{|\mathcal D_{\mathrm{cal}}|}
 \sum_{(\bm x,y)\in\mathcal D_{\mathrm{cal}}}
 \ell_{\mathrm{CE}}\bigl(f_{\bm{\theta}(\bm{\lambda})}(\bm x),y\bigr),
 \label{eq:cal_ce}
\end{equation}
where $f_{\bm{\theta}}$ is the classifier with parameters $\bm{\theta}$,
$(\bm x,y)$ is an image--label pair, $\ell_{\mathrm{CE}}$ is the cross-entropy loss,
and $|\mathcal D_{\mathrm{cal}}|$ is the number of calibration examples.
FedQCal optimizes
\begin{equation}
 \min_{\bm a\in\mathbb{R}_+^2}\;L_{\mathrm{cal}}\bigl(\bm{\lambda}(\bm a)\bigr),
 \label{eq:dual_ce}
\end{equation}
where $\mathbb{R}_+=[0,\infty)$ includes zero.
Each candidate $(u,v)$ specifies coefficients through Eq.~\eqref{eq:coefficient_map}
and a merged model through Eq.~\eqref{eq:merge}.
Backpropagation through the complete trainable block
and classifier updates the coordinates using the actual merged network,
without linearizing its output or approximating CE.
As illustrated on the left of Figure~\ref{fig:geometry}(a),
only two coordinates are optimized, regardless of the number of clients.
When both score sums are positive, we initialize the coordinates as
\begin{equation}
 u_0=\frac{1}{2\sum_k\widehat{\gamma}_k},
 \quad
 v_0=\frac{1}{2\sum_k\widetilde g_k}.
 \label{eq:initialization}
\end{equation}
Before clipping, this gives an equal mixture of the two
sum-normalized score vectors.
The original score scales are retained during optimization,
and the coefficients are not constrained to sum to one.

To optimize Eq.~\eqref{eq:dual_ce},
FedQCal uses projected gradient directions with Armijo backtracking~\cite{armijo1966minimization}.
Projection keeps the two coordinates nonnegative,
while backtracking selects a step that sufficiently decreases calibration loss.
Appendix~\ref{app:optimization} gives the gradient and backtracking condition.

The resulting coefficients define the final classifier through Eq.~\eqref{eq:merge}.
Calibration runs entirely on the server with fixed client updates and alignment scores,
so no further client communication is required.
Algorithm~\ref{alg:fedqp} summarizes the procedure, including its FedAvg fallback
when no reference-alignment signal is available.

\section{QP Structure and Reference Alignment}
\label{sec:analysis}
We characterize calibrated coefficients as QP solutions
and analyze their reference projection.
Throughout, let $\bm a\in\mathbb{R}_+^2$, $\mathcal B=[-b,b]^K$,
and $\bm{\lambda}(\bm a)=\Pi_{\mathcal B}(\bm C^\top\bm a)$,
where $\Pi_{\mathcal B}$ denotes Euclidean projection onto the box (componentwise clipping)
and $\bm C=[\hat{\bm\gamma}^{\top};\tilde{\bm g}^{\top}]$ is fixed after reference training.

\subsection{Every calibrated point solves an induced QP}
\label{sec:qp_characterization}
\begin{proposition}[QP characterization]
\label{prop:implicit}
For any $\bm a\ge\bm0$, define the induced thresholds $\bm m(\bm a)=\bm C\bm{\lambda}(\bm a)$.
Then $\bm{\lambda}(\bm a)$ is the unique solution of
\begin{equation}
\min_{\bm x\in\mathcal B}\ \tfrac12\|\bm x\|_2^2
\quad\text{s.t.}\quad
\bm C\bm x\ge\bm m(\bm a).
\label{eq:implicit_qp}
\end{equation}
Conversely, the optimizer of every feasible fixed-threshold QP in Eq.~\eqref{eq:qp} has this
form for some $\bm a\ge\bm0$.
\end{proposition}

The key observation is that box projection minimizes $\frac12\|\bm x\|_2^2-\bm a^\top\bm C\bm x$ over $\mathcal B$.
Any point meeting the induced thresholds has at least the same weighted score,
so it cannot have a smaller norm than $\bm{\lambda}(\bm a)$.
The converse follows from the Karush--Kuhn--Tucker (KKT) optimality conditions for linear
constraints, including boundary-feasible thresholds; full proofs are in Appendix~\ref{app:implicit}.
Thus calibration selects the alignment requirements together with the coefficients.
At the selected requirements, the resulting coefficients remain the unique minimum-norm
solution of an inequality-constrained QP.
The thresholds are outputs of this selection,
rather than prescribed values that CE must preserve.

\paragraph{A certificate at every calibration step.}
The QP guarantee does not require CE optimization to converge.
For any nonnegative coordinates, the induced coefficients lie in the box
and attain both induced score thresholds exactly.
The same coefficients minimize the box-constrained Lagrangian at those coordinates.
Consequently, the QP objective equals the lower bound given by its dual,
$\tfrac12\|\bm{\lambda}(\bm a)\|_2^2$.
This gives an optimality certificate from the scores, coordinates,
and coefficients, without solving another QP.
A finite calibration run therefore retains the minimum-norm property
for the requirements reached at its final step.

\subsection{Calibration over the fixed-threshold solution family}
\label{sec:family_scope}
Each calibrated point solves an induced QP.
We next examine which fixed-threshold QP solutions calibration can select.
Proposition~\ref{prop:implicit} shows that the two-coordinate parameterization
includes every solution obtained by fixing feasible alignment requirements.
Calibration can therefore use classification loss to choose among all such QP solutions.
In particular, for any feasible fixed-threshold solution $\bm{\lambda}^{\mathrm{fix}}$,
\begin{equation}
\inf_{\bm a\ge\bm0}L_{\mathrm{cal}}\bigl(\bm{\lambda}(\bm a)\bigr)
\le L_{\mathrm{cal}}(\bm{\lambda}^{\mathrm{fix}}),
\label{eq:calibration_order}
\end{equation}
where $\inf$ denotes the greatest lower bound over the stated domain.

Equation~\eqref{eq:calibration_order} compares a global infimum with a feasible candidate.
The actual CE objective is generally nonconvex,
so finite optimization need not attain this infimum or
improve on a particular fixed-threshold run.
Nor does an ordering of calibration losses imply an ordering of test accuracies.
The experiments evaluate the benefit of the implemented calibration procedure;
the proposition establishes its QP structure and representational scope.

The comparison in Eq.~\eqref{eq:calibration_order} applies only to the QP solution family
determined by the alignment scores.
Appendix~\ref{app:ce_search} describes the resulting restrictions on the client coefficients.

\subsection{Coefficient norm and projection sensitivity}
\label{sec:perturb}
The alignment constraints alone can admit many coefficient vectors.
FedQCal selects the one with minimum norm.
We now explain this choice when the reference and client summaries are imperfect.
The selected coefficients determine how strongly errors in these matrices
can affect the aggregate reference projection.
The analysis explains the minimum-norm objective as a way to meet
the selected alignment requirements while limiting the amplification of projection errors.

For $\Delta\bm W_{\mathrm{ref}}\ne\bm0$, let
$\widehat{\bm R}=\Delta\bm W_{\mathrm{ref}}/\|\Delta\bm W_{\mathrm{ref}}\|_F$
be the measured reference direction.
Equations~\eqref{eq:rho}--\eqref{eq:scores} give
$\tilde g_k=\max\{0,\left\langle \Delta\bm W_k,\widehat{\bm R}\right\rangle_F\}$.
A client with $\left\langle \Delta\bm W_k,\widehat{\bm R}\right\rangle_F\le0$ has
both scores zero and hence zero coefficient.
For all remaining clients, the magnitude score equals their reference projection.
Consequently,
\begin{equation}
\left\langle \sum_k\lambda_k(\bm a)\Delta\bm W_k,\widehat{\bm R}\right\rangle_F
=\tilde{\bm g}^\top\bm{\lambda}(\bm a).
\label{eq:aggregate_projection}
\end{equation}
Thus the second QP requirement has a direct geometric meaning:
it controls the projection of the weighted sum of client summaries along the reference direction.
For the selected coefficients, this projection equals $m_2(\bm a)$.

Equation~\eqref{eq:aggregate_projection} uses the measured summaries and reference.
To assess how errors in these matrices affect the interpretation of this projection,
let $\Delta\bm W_k^*$ and $\Delta\bm W_{\mathrm{ref}}^*\ne\bm0$ denote hypothetical unperturbed counterparts.
Suppose they satisfy $\|\Delta\bm W_k-\Delta\bm W_k^*\|_F\le\epsilon_k$
and $\|\Delta\bm W_{\mathrm{ref}}-\Delta\bm W_{\mathrm{ref}}^*\|_F\le\epsilon_{\mathrm{ref}}<\|\Delta\bm W_{\mathrm{ref}}^*\|_F$.
Here $\Delta\bm W_{\mathrm{ref}}^*/\|\Delta\bm W_{\mathrm{ref}}^*\|_F$ is the ideal reference direction
normalized to unit Frobenius norm.
Define the error vector $\bm e=(e_1,\ldots,e_K)^\top$
by $e_k=\epsilon_k+2\|\Delta\bm W_k^*\|_F\epsilon_{\mathrm{ref}}/\|\Delta\bm W_{\mathrm{ref}}^*\|_F$.
Each $e_k$ bounds the resulting error in a client's projection.
\begin{samepage}
\begin{proposition}[Target-projection bound]
\label{prop:perturb}
For any $\bm a\ge\bm0$ and induced coefficients $\bm{\lambda}=\bm{\lambda}(\bm a)$,
\begin{align}
&\left\langle \sum_k\lambda_k\Delta\bm W_k^*,\frac{\Delta\bm W_{\mathrm{ref}}^*}{\|\Delta\bm W_{\mathrm{ref}}^*\|_F}\right\rangle_F\nonumber\\
&\ge\tilde{\bm g}^\top\bm{\lambda}-\sum_k\lambda_ke_k\nonumber\\
&\ge\tilde{\bm g}^\top\bm{\lambda}-\|\bm{\lambda}\|_2\|\bm e\|_2.
\label{eq:projection_bound}
\end{align}
\end{proposition}
\end{samepage}

For the selected coefficients, Eq.~\eqref{eq:projection_bound} gives a lower bound
on the projection of the unperturbed summaries along the ideal reference direction:
the measured projection minus an allowance for uncertainty.
For a common bound on $\|\bm e\|_2$,
the norm-based allowance grows in proportion to $\|\bm{\lambda}\|_2$.
Figure~\ref{fig:geometry}(b) illustrates how merges with the same aggregate alignment scores
can differ in sensitivity to reference error.
By Proposition~\ref{prop:implicit}, the QP solution minimizes this allowance
among coefficients satisfying the selected alignment requirements.
The constraints retain the required alignment,
while the norm objective limits the amplification of projection errors.
CE also changes the selected requirements, so this argument does not imply
that calibration maximizes the lower bound as a whole.

\paragraph{Interpreting the uncertainty allowance.}
The relative error $\epsilon_{\mathrm{ref}}/\|\Delta\bm W_{\mathrm{ref}}^*\|_F$ governs reference sensitivity:
a smaller reference permits greater directional uncertainty for a fixed absolute error.
The bound uses coefficient norm; the merged parameter update norm also depends
on correlations among client task vectors.

The bound concerns the weighted attention summaries in Eq.~\eqref{eq:aggregate_projection}.
The actual merged key--query product additionally contains cross-client terms,
and the bound does not guarantee improved classification accuracy.
Appendix~\ref{app:projection_proof} gives the proof, and
Appendix~\ref{app:geometry} derives and measures this discrepancy.
\begin{table*}[t]
\centering
\caption{\textbf{One-shot federated learning on CIFAR-10, CIFAR-100 and SVHN.}
Test accuracy (\%), mean $\pm$ SD over three seeds.
Server-supervised methods use five calibration images per class.
Bold and underlined values indicate the highest and second-highest mean accuracies,
respectively, within each client-count block and column.}
\label{tab:fedqp_main}
\small
\setlength{\tabcolsep}{1.5pt}
\medmuskip=1mu
\resizebox{\textwidth}{!}{%
\begin{tabular}{lccccccccc}
\toprule
& \multicolumn{3}{c}{CIFAR-10}
& \multicolumn{3}{c}{CIFAR-100}
& \multicolumn{3}{c}{SVHN}\\
\cmidrule(lr){2-4}\cmidrule(lr){5-7}\cmidrule(lr){8-10}
Method
& $\alpha=0.05$ & $\alpha=0.10$ & $\alpha=0.50$
& $\alpha=0.05$ & $\alpha=0.10$ & $\alpha=0.50$
& $\alpha=0.05$ & $\alpha=0.10$ & $\alpha=0.50$\\
\midrule

\multicolumn{10}{l}{\emph{10 clients}}\\

Uniform averaging
& $59.66\pm11.00$ & $72.33\pm16.68$ & $91.73\pm0.09$
& $37.26\pm2.04$ & $47.96\pm2.12$ & $69.97\pm0.64$
& $20.23\pm5.51$ & $19.82\pm2.38$ & $45.60\pm1.46$\\

FedAvg
& $71.46\pm3.60$ & $76.21\pm5.27$ & $\underline{91.92\pm0.39}$
& $37.84\pm2.13$ & $48.45\pm1.20$ & $69.95\pm0.60$
& $21.67\pm2.67$ & $23.72\pm4.37$ & $46.78\pm1.30$\\

Ensemble
& $61.07\pm12.05$ & $72.14\pm17.45$ & $91.82\pm0.18$
& $38.20\pm1.77$ & $48.77\pm2.28$ & $\underline{70.35\pm0.44}$
& $22.88\pm8.74$ & $25.83\pm5.91$ & $\underline{49.27\pm0.88}$\\

TIES-Merging
& $58.61\pm11.12$ & $70.88\pm18.95$ & $91.87\pm0.09$
& $37.30\pm2.14$ & $49.00\pm2.14$ & $70.26\pm0.62$
& $20.28\pm5.85$ & $18.97\pm1.70$ & $44.32\pm3.23$\\

\cmidrule(lr){1-10}

Reference-only
& $70.01\pm2.83$ & $70.01\pm2.83$ & $70.01\pm2.83$
& $51.47\pm1.28$ & $51.47\pm1.28$ & $51.47\pm1.28$
& $16.50\pm0.14$ & $16.50\pm0.14$ & $16.50\pm0.14$\\

FedAvg + calibration
& $\mathbf{86.06\pm0.73}$
& $\mathbf{87.59\pm1.24}$
& $90.31\pm0.54$
& $\mathbf{62.30\pm0.97}$
& $\mathbf{64.18\pm0.87}$
& $69.27\pm0.85$
& $\underline{25.68\pm1.91}$
& $\underline{28.95\pm1.42}$
& $36.41\pm1.49$\\

FedQCal (ours)
& $\underline{78.08\pm1.39}$
& $\underline{85.07\pm5.38}$
& $\mathbf{92.20\pm0.26}$
& $\underline{52.03\pm2.57}$
& $\underline{58.46\pm3.39}$
& $\mathbf{71.83\pm0.61}$
& $\mathbf{39.37\pm5.75}$
& $\mathbf{41.39\pm5.12}$
& $\mathbf{52.52\pm1.31}$\\

\midrule
\midrule
\multicolumn{10}{l}{\emph{100 clients}}\\

Uniform averaging
& $54.47\pm6.06$ & $80.90\pm1.95$ & $90.91\pm0.33$
& $20.95\pm1.57$ & $36.63\pm0.40$ & $62.62\pm0.30$
& $16.59\pm1.85$ & $26.47\pm2.27$ & $45.17\pm0.76$\\

FedAvg
& $76.96\pm3.34$ & $84.46\pm2.42$ & $\underline{91.03\pm0.30}$
& $22.56\pm1.94$ & $38.34\pm0.35$ & $62.95\pm0.29$
& $18.57\pm4.59$ & $30.03\pm3.86$ & $46.72\pm0.36$\\

Ensemble
& $56.82\pm7.03$ & $82.88\pm2.08$ & $\mathbf{91.54\pm0.27}$
& $22.32\pm1.75$ & $38.81\pm0.36$ & $64.07\pm0.15$
& $21.23\pm2.99$ & $\underline{34.02\pm4.28}$ & $\mathbf{51.07\pm1.29}$\\

TIES-Merging
& $47.07\pm3.29$ & $78.19\pm2.52$ & $90.95\pm0.20$
& $24.11\pm2.11$ & $40.83\pm0.65$ & $\underline{65.56\pm0.36}$
& $17.81\pm1.76$ & $27.67\pm4.71$ & $43.98\pm0.60$\\

\cmidrule(lr){1-10}

Reference-only
& $70.14\pm3.14$ & $70.14\pm3.14$ & $70.14\pm3.14$
& $\underline{51.21\pm0.91}$ & $51.21\pm0.91$ & $51.21\pm0.91$
& $17.03\pm0.30$ & $17.03\pm0.30$ & $17.03\pm0.30$\\

FedAvg + calibration
& $\underline{85.38\pm0.34}$
& $\underline{86.39\pm0.42}$
& $89.11\pm0.24$
& $\mathbf{59.11\pm1.13}$
& $\mathbf{60.55\pm1.25}$
& $63.83\pm0.99$
& $\underline{23.85\pm1.10}$
& $26.41\pm1.20$
& $33.18\pm2.04$\\

FedQCal (ours)
& $\mathbf{89.49\pm0.62}$
& $\mathbf{88.49\pm0.13}$
& $90.51\pm1.59$
& $51.12\pm1.61$
& $\underline{57.30\pm0.68}$
& $\mathbf{66.26\pm0.08}$
& $\mathbf{37.91\pm3.44}$
& $\mathbf{43.72\pm1.95}$
& $\underline{47.16\pm1.97}$\\

\bottomrule
\end{tabular}%
}
\end{table*}

\section{Experiments}
\label{sec:experiments}
The main comparison evaluates FedQCal against existing aggregation methods across label skew,
datasets, and client counts.

\subsection{Experimental setup}
\label{sec:protocol}
\paragraph{Data and evaluation.}
We evaluate CIFAR-10, CIFAR-100~\cite{krizhevsky2009learning}, and SVHN~\cite{svhn} with 10 or 100 clients
and Dirichlet label-skew parameters $\alpha\in\{0.05,0.10,0.50\}$.
For SVHN, we use the standard training and test splits without the extra split.
Results are mean test accuracy and sample standard deviation over
training seeds 42, 43, and 44, using the complete test sets.
Before client partitioning, five training images per class are held
out for server use.
The main comparison uses all five images, while the one-image-per-class
setting selects a subset of the same holdout.
Unused held-out images remain excluded from client training.
Reference training and coefficient calibration use the same selected
server pool.

\paragraph{Model and training.}
All methods use the same pretrained ViT-Tiny/16 checkpoint
(ImageNet-21k pretraining followed by ImageNet-1k fine-tuning)
as the initialization for training on CIFAR-10, CIFAR-100, and SVHN.
All inputs have a resolution of $224\times224$.
Client and reference training update only the final Transformer block
and classifier, leaving the other model parameters frozen.
Client and reference models use the same training settings:
SGD with learning rate $0.001$, momentum $0.9$, weight decay
$5\times10^{-4}$, and batch size 64, for 30 and 300 epochs with
10 and 100 clients, respectively.
Reference training draws a fixed sample list from the selected server
pool, sized to the mean client sample count rounded down, with
replacement only when required.

\paragraph{Baselines.}
Uniform averaging assigns equal weights to client models, whereas
one-round FedAvg~\cite{mcmahan2017communication} weights them by local sample counts.
Ensemble uniformly averages client logits and retains all client
models for inference.
For TIES-Merging~\cite{yadav2024ties}, we retain the largest 20\%
of absolute entries in each client task vector and perform sign election.
We uniformly average the processed vectors over all clients,
including zeroed entries, with the merge scale fixed to one.
No labeled server data are used.
Reference-only evaluates the reference model trained from the shared
initialization, without aggregating client updates.
FedAvg + calibration directly optimizes the final Transformer block
and classifier from sample-weighted FedAvg using calibration CE and
Armijo backtracking.
Methods that aggregate the trained client models reuse the same
saved checkpoints.

\paragraph{FedQCal configuration.}
FedQCal uses raw attention-update summaries, normalized-score
initialization, two nonnegative coordinates, and coefficient cap $b=2$.
All CE-based calibration variants minimize mean CE over the complete
selected server pool with deterministic preprocessing and the model
in evaluation mode.
They report the last accepted model within at most 100 accepted
Armijo updates.
Test accuracy is used only for evaluation, not endpoint selection.
Implementation details and fallback rules appear in
Appendix~\ref{sec:implementation}.

\begin{table*}[t]
\centering
\caption{\textbf{Ablation results with 10 clients.}
Test accuracy (\%), mean $\pm$ SD over three training seeds.
$n$ denotes the number of labeled server images per class.
The first four rows are FedQCal variants; random directions average five pairs per seed.}
\label{tab:fedqp_ablations}
\small
\setlength{\tabcolsep}{3.2pt}
\medmuskip=1mu
\resizebox{\textwidth}{!}{%
\begin{tabular}{lccccccccc}
\toprule
& \multicolumn{3}{c}{CIFAR-10}
& \multicolumn{3}{c}{CIFAR-100}
& \multicolumn{3}{c}{SVHN}\\
\cmidrule(lr){2-4}\cmidrule(lr){5-7}\cmidrule(lr){8-10}
Setting
& $\alpha=0.05$ & $\alpha=0.10$ & $\alpha=0.50$
& $\alpha=0.05$ & $\alpha=0.10$ & $\alpha=0.50$
& $\alpha=0.05$ & $\alpha=0.10$ & $\alpha=0.50$\\
\midrule

\multicolumn{10}{l}{\emph{Five server images per class ($n=5$)}}\\

FedQCal
& $78.08\pm1.39$ & $85.07\pm5.38$ & $92.20\pm0.26$
& $52.03\pm2.57$ & $58.46\pm3.39$ & $71.83\pm0.61$
& $39.37\pm5.75$ & $41.39\pm5.12$ & $52.52\pm1.31$\\

Initialization only
& $69.94\pm1.56$ & $78.60\pm11.35$ & $91.87\pm0.13$
& $37.81\pm2.33$ & $48.42\pm1.97$ & $70.07\pm0.67$
& $23.87\pm4.84$ & $21.93\pm1.93$ & $46.44\pm0.72$\\

Random directions
& $71.75\pm3.72$ & $80.94\pm2.68$ & $91.60\pm0.63$
& $47.11\pm2.48$ & $53.99\pm4.31$ & $70.16\pm0.30$
& $32.66\pm8.84$ & $38.16\pm2.74$ & $51.15\pm0.46$\\

Independent head
& $85.52\pm0.78$ & $87.07\pm1.07$ & $90.44\pm0.61$
& $62.47\pm1.03$ & $64.09\pm0.82$ & $69.04\pm0.96$
& $23.90\pm1.62$ & $26.71\pm2.52$ & $34.73\pm1.79$\\

FedAvg + calibration
& $86.06\pm0.73$ & $87.59\pm1.24$ & $90.31\pm0.54$
& $62.30\pm0.97$ & $64.18\pm0.87$ & $69.27\pm0.85$
& $25.68\pm1.91$ & $28.95\pm1.42$ & $36.41\pm1.49$\\

\midrule
\multicolumn{10}{l}{\emph{One server image per class ($n=1$)}}\\

FedQCal
& $74.93\pm5.08$ & $85.58\pm3.01$ & $92.05\pm0.27$
& $51.31\pm3.53$ & $58.10\pm4.04$ & $71.63\pm0.74$
& $39.01\pm5.48$ & $43.46\pm3.32$ & $52.92\pm0.24$\\

Initialization only
& $69.72\pm1.50$ & $78.34\pm10.61$ & $91.90\pm0.13$
& $37.68\pm2.37$ & $48.42\pm1.99$ & $70.06\pm0.64$
& $23.58\pm4.98$ & $21.40\pm0.72$ & $46.57\pm0.62$\\

Random directions
& $71.96\pm6.85$ & $80.71\pm3.97$ & $91.42\pm0.44$
& $45.56\pm4.19$ & $53.95\pm4.39$ & $69.92\pm0.51$
& $34.24\pm5.60$ & $38.80\pm6.21$ & $50.77\pm1.31$\\

Independent head
& $81.07\pm4.59$ & $84.67\pm2.33$ & $90.98\pm0.64$
& $49.17\pm1.28$ & $55.30\pm0.46$ & $67.12\pm0.64$
& $19.09\pm3.40$ & $23.59\pm2.47$ & $36.55\pm3.97$\\

FedAvg + calibration
& $81.59\pm4.34$ & $84.35\pm1.16$ & $90.80\pm0.80$
& $48.54\pm1.37$ & $54.90\pm0.44$ & $67.13\pm0.72$
& $20.81\pm3.81$ & $25.54\pm3.94$ & $35.75\pm4.48$\\

\bottomrule
\end{tabular}%
}
\end{table*}

\subsection{Comparison with existing methods}

FedQCal achieves higher mean accuracy than FedAvg in 17 of the 18
settings in Table~\ref{tab:fedqp_main}.
Relative to FedAvg + calibration, it performs better in five of nine
settings with 10 clients and seven of nine settings with 100 clients.
Its relative performance is therefore more favorable in the 100-client
setting under our protocol.
This trend remains dataset-dependent: FedQCal outperforms direct
calibration in all six SVHN settings, whereas on CIFAR-100 with
five server images per class, its advantage is confined to
\(\alpha=0.50\) with both client counts.

On SVHN, Reference-only achieves about 17\% accuracy,
yet FedQCal outperforms it and FedAvg in all six settings.
This suggests that a weak classifier can still serve
as a useful aggregation reference.
Appendix~\ref{sec:additional_comparisons} extends these comparisons to ViT-S
and one labeled server image per class.

\subsection{Which components matter?}

Table~\ref{tab:fedqp_ablations} presents ablations with 10 clients.
Appendix~\ref{sec:additional_ablations} extends them and examines average
sensitivity to the server-data budget.

\paragraph{Effect of coefficient calibration.}
FedQCal (initialization only) retains the normalized-score initial
coefficients without CE-based coefficient optimization.
Comparing this variant with FedQCal therefore tests the contribution
of calibration while keeping the reference model, alignment scores,
and coefficient map unchanged within each budget.
FedQCal achieves higher mean accuracy in all 18 settings.
These gains support the empirical value of CE-based selection within
the QP solution family.

\paragraph{Reference-guided weighting.}
To assess whether the reference-guided scores contribute beyond a
two-variable parameterization, FedQCal (random directions) replaces
$\hat{\bm\gamma}$ and $\tilde{\bm g}$ with independent directions sampled
uniformly from the nonnegative unit sphere and scaled to match the
corresponding score-vector $\ell_2$ norms.
Both variants optimize two nonnegative coordinates using the same
calibration data, coefficient cap, and Armijo protocol.
The random control follows the same reciprocal-sum initialization rule,
matching the initial coefficient sum while generally producing a
different initial merged model.

FedQCal achieves higher mean accuracy in all 18 settings in
Table~\ref{tab:fedqp_ablations}, with gains of 0.59--6.71 percentage points
over the random-direction control.
Within each dataset and server-data budget, the gains are larger at
$\alpha=0.05$ and $0.10$ than at $\alpha=0.50$.
These results support the reference-guided weighting relative to the
tested random two-direction control at the same optimization dimension.

\paragraph{Independent classifier adaptation.}
FedQCal (independent head) starts from the same initial merged model
as standard FedQCal and jointly optimizes the two coordinates and
all classifier weights and biases, while retaining the original
Transformer-block parameterization.
On CIFAR-100, independent classifier adaptation improves mean accuracy
at the two smaller values of $\alpha$ with five server images per class.
With one image per class, standard FedQCal performs better
at all three values of $\alpha$ in Table~\ref{tab:fedqp_ablations}.

The independent-head variant and FedAvg + calibration obtain
comparable mean accuracies across all 18 settings, with absolute
differences no larger than 2.25 percentage points, despite their
different initializations and Transformer-block parameterizations.
Although the independent-head variant achieves lower calibration CE
than standard FedQCal in every run, its additional flexibility does
not consistently improve test accuracy.
Together, these observations suggest that coupling the classifier
and Transformer block through shared merge coefficients can provide
a useful constraint when labeled server data are limited.

\paragraph{Server-data budget.}
Finally, reducing the budget from five images per class to one
changes FedQCal's mean accuracy by less than one percentage point
in seven of the nine dataset--skew settings.
The largest decrease occurs on CIFAR-10 at $\alpha=0.05$,
where accuracy decreases from 78.08\% to 74.93\%.
Additional images do not uniformly improve accuracy.
Because the budget changes both reference training and coefficient
calibration, this comparison evaluates its effect on FedQCal
as a whole.
Compared with independent head and FedAvg + calibration,
FedQCal's lower average budget sensitivity supports
constrained calibration with scarce server labels.

\subsection{Scope and limitations}
\label{sec:limits}
FedQCal calibrates a merge family determined by fixed client updates
and reference-alignment scores.
Within this family, clients with nonpositive reference alignment
retain zero coefficients throughout calibration, so CE optimization
cannot restore contributions excluded by the reference scores.
Although only two coordinates are optimized, coefficient fitting
requires forward evaluation of the merged network and backpropagation
through the coefficient-dependent layers.
Server computation also includes reference training.

\section{Conclusion}

FedQCal uses limited server supervision to guide one-shot aggregation
of heterogeneous client updates.
A reference adaptation supplies a common task-based criterion
for comparing these updates.
Classification loss then selects the required degree of alignment
within a two-coordinate family of minimum-norm QP solutions.
This formulation preserves the QP property throughout calibration.
The projection analysis explains how the minimum-norm objective
limits the amplification of projection errors
at fixed alignment requirements.

Experiments on CIFAR-10, CIFAR-100, and SVHN show gains over
one-round FedAvg in most settings.
The ablations support both reference-guided weighting
and coefficient calibration,
even with one labeled server example per class.
The gains over the tested random-direction control are larger
under stronger label skew.
Together, these results support a design principle:
limited server supervision can define a task-informed family of merges
and select a useful model within it.
FedQCal realizes this principle with two calibration variables
and an explicit minimum-norm characterization.

{\small
\pdfbookmark[1]{References}{references}
\bibliographystyle{ieeenat_fullname}
\bibliography{references}
}

\twocolumn[{%
  \pdfbookmark[0]{Supplementary Material}{supplementary}%
  \centering
  {\Large\bfseries Supplementary Material\par}
  \vspace{1em}
}]
\appendix
\setcounter{equation}{0}
\setcounter{table}{0}
\setcounter{figure}{0}
\setcounter{proposition}{0}
\renewcommand{\theequation}{S\arabic{equation}}
\renewcommand{\thetable}{S\arabic{table}}
\renewcommand{\thefigure}{S\arabic{figure}}
\renewcommand{\theproposition}{S\arabic{proposition}}
\renewcommand{\theHequation}{supp.\arabic{equation}}
\renewcommand{\theHtable}{supp.\arabic{table}}
\renewcommand{\theHfigure}{supp.\arabic{figure}}

\section{Dual Coordinates and Quadratic Programs}
\label{app:proofs}
\subsection{Notation and the fixed-threshold problem}
Let $K$ be the number of clients.
The updates $\Delta\bm W_k$ and $\Delta\bm W_{\mathrm{ref}}$ describe changes in attention summaries
from the shared initialization for client $k$ and the reference, respectively.
For $\Delta\bm W_{\mathrm{ref}}\ne\bm0$, define the normalized reference direction by
$\widehat{\bm R}=\Delta\bm W_{\mathrm{ref}}/\|\Delta\bm W_{\mathrm{ref}}\|_F$.
Define
\begin{align}
h_k&=\left\langle \Delta\bm W_k,\widehat{\bm R}\right\rangle_F,
&\tilde g_k&=\max\{h_k,0\},\\
\hat\gamma_k&=\sqrt{\tilde g_k/\|\Delta\bm W_k\|_F},
&\bm C&=[\hat{\bm\gamma}^{\top};\tilde{\bm g}^{\top}],
\end{align}
where $\left\langle \bm X,\bm Y\right\rangle_F=\sum_{i,j}X_{ij}Y_{ij}$ is the Frobenius inner product
and $\|\bm X\|_F=\sqrt{\left\langle \bm X,\bm X\right\rangle_F}$ is the Frobenius norm.
The superscript $\top$ denotes transpose, and $\hat{\bm\gamma}$
and $\tilde{\bm g}$ collect the corresponding scores over all clients.
In finite-precision computation, negative cosine similarity values are set to zero
before taking the square root.
A zero client summary receives both scores zero.
Let $b>0$ be the coefficient magnitude cap, $\mathcal B=[-b,b]^K$ the coefficient box,
and $\bm m=(m_1,m_2)^\top$ the required score thresholds.
The fixed-threshold FedQCal quadratic program (QP) is
\begin{equation}
\min_{\bm x\in\mathcal B}\ \tfrac12\|\bm x\|_2^2
\quad\text{s.t.}\quad\bm C\bm x\ge\bm m,
\label{eq:s_qp}
\end{equation}
where $\|\bm x\|_2=\sqrt{\sum_k x_k^2}$ is the Euclidean norm.
Vector and matrix inequalities are interpreted componentwise.
Because $\bm C\ge\bm0$, this problem is feasible exactly when $m_1\le b\sum_k\hat\gamma_k$
and $m_2\le b\sum_k\tilde g_k$.
Indeed, the point $b\bm1$, where $\bm1$ is the all-ones vector,
simultaneously maximizes both constraint sums over $\mathcal B$.
A nonempty feasible set is compact and convex, and the objective is strictly convex;
hence the optimizer exists and is unique.
Replacing a negative coordinate by zero cannot reduce either score
and strictly reduces its squared norm.
Thus the optimizer is nonnegative.

\subsection{Dual objective and coefficient map}
\label{app:duality}
For the nonnegative constraint multipliers $\bm a=(u,v)^\top\ge\bm0$,
the box-restricted Lagrangian is
\begin{equation}
L(\bm x,\bm a;\bm m)
=\tfrac12\|\bm x\|_2^2+\bm a^\top(\bm m-\bm C\bm x).
\end{equation}
Its minimizer over $\mathcal B$ is
\begin{equation}
\bm{\lambda}(\bm a)=\Pi_{\mathcal B}(\bm C^\top\bm a),\quad
\lambda_k(\bm a)=\operatorname{clip}
 (u\hat\gamma_k+v\tilde g_k,-b,b),
\label{eq:s_coefficient_map}
\end{equation}
where $\Pi_{\mathcal S}$ denotes Euclidean projection onto a nonempty closed convex set
$\mathcal S$: it returns the point in $\mathcal S$ closest
to its argument in Euclidean distance.
The clipping operation is $\operatorname{clip}(s,-b,b)=\min\{b,\max\{-b,s\}\}$.
The lower clipping bound is inactive for nonnegative coordinates.
Writing
\begin{equation}
\psi_b(s)=
\begin{cases}
\tfrac12s^2,&|s|\le b,\\
b|s|-\tfrac12b^2,&|s|>b,
\end{cases}
\end{equation}
the concave QP dual objective is
\begin{equation}
q(\bm a;\bm m)=\bm a^\top\bm m
-\sum_{k=1}^K\psi_b\big((\bm C^\top\bm a)_k\big),
\quad\bm a\ge\bm0.
\label{eq:s_dual}
\end{equation}
It is continuously differentiable, with $\nabla_{\bm a}q=\bm m-\bm C\bm{\lambda}(\bm a)$.
Here $\nabla_{\bm a}$ denotes the gradient with respect to $\bm a$.
For prescribed thresholds, the multipliers maximize this objective
and obey complementary slackness: $a_j((\bm C\bm{\lambda})_j-m_j)=0$ for $j=1,2$.
FedQCal instead minimizes the calibration cross-entropy (CE) of the merged model through
Eq.~\eqref{eq:s_coefficient_map}.
These are different optimization objectives.

\subsection{Proof of the implicit-threshold characterization}
\label{app:implicit}
Fix any $\bm a\ge\bm0$, let $\bm{\lambda}=\bm{\lambda}(\bm a)$, and set $\bm m=\bm C\bm{\lambda}$.
The projection optimality condition is
\begin{equation}
\left\langle \bm{\lambda}-\bm C^\top\bm a,\bm x-\bm{\lambda}\right\rangle\ge0
\quad\text{for all }\bm x\in\mathcal B,
\label{eq:s_projection_optimality}
\end{equation}
where $\left\langle \bm x,\bm y\right\rangle=\bm x^\top\bm y$ is the Euclidean inner product.
For any $\bm x\in\mathcal B$ satisfying $\bm C\bm x\ge\bm m$,
\begin{align}
\tfrac12\|\bm x\|_2^2-\tfrac12\|\bm{\lambda}\|_2^2
&=\left\langle \bm{\lambda},\bm x-\bm{\lambda}\right\rangle+\tfrac12\|\bm x-\bm{\lambda}\|_2^2\nonumber\\
&\ge\bm a^\top\bm C(\bm x-\bm{\lambda})+\tfrac12\|\bm x-\bm{\lambda}\|_2^2\nonumber\\
&\ge\tfrac12\|\bm x-\bm{\lambda}\|_2^2.
\label{eq:s_implicit_proof}
\end{align}
This proves that $\bm{\lambda}$ is the unique optimizer of
Eq.~\eqref{eq:s_qp} at these induced thresholds.
It is also an explicit primal--dual certificate: feasibility holds by construction,
$\bm a$ is dual-feasible, and both score constraints are tight.
No strict-feasibility assumption is required.

Conversely, let $\bm{\lambda}^*$ solve a feasible instance of Eq.~\eqref{eq:s_qp}.
The normal cone of its polyhedral feasible set is generated by
the normals of the active linear constraints.
First-order optimality therefore gives nonnegative multipliers $\bm a$ such that
\begin{equation}
\bm0\in\bm{\lambda}^*-\bm C^\top\bm a+N_{\mathcal B}(\bm{\lambda}^*),
\end{equation}
with the usual score-constraint complementarity conditions.
For $\bm z\in\mathcal B$, the normal cone of the box is
\begin{equation}
N_{\mathcal B}(\bm z)
=\{\bm r:\left\langle \bm r,\bm x-\bm z\right\rangle\le0\text{ for all }\bm x\in\mathcal B\}.
\end{equation}
The stationarity condition is equivalent to Eq.~\eqref{eq:s_projection_optimality},
so $\bm{\lambda}^*=\Pi_{\mathcal B}(\bm C^\top\bm a)$.
The polyhedral normal-cone argument also covers feasible threshold pairs on the boundary,
where a strictly feasible point, satisfying every inequality strictly, need not exist.
The coefficient vector is unique; its multiplier representation need not be.

The characterization shows that the FedQCal family consists of QP optimizers with
adaptively selected score requirements.
It does not assert that CE optimization increases both scores over the fixed-threshold QP solution, or that it preserves the thresholds of that QP.

\subsection{Calibration objective and coefficient family}
\label{app:ce_search}
Let $\bm{\theta}_0$ and $\bm{\theta}_k$ denote the shared initial and client parameters, respectively,
and let $\mathcal D_{\mathrm{cal}}$ be the labeled server set,
with size $|\mathcal D_{\mathrm{cal}}|$.
The calibrated model uses the original client task vectors $\Delta\bm{\theta}_k=\bm{\theta}_k-\bm{\theta}_0$:
\begin{equation}
\begin{aligned}
\bm{\theta}(\bm{\lambda})&=\bm{\theta}_0+\sum_k\lambda_k\Delta\bm{\theta}_k,\\
L_{\mathrm{cal}}(\bm{\lambda})
&=\frac1{|\mathcal D_{\mathrm{cal}}|}
\sum_{(\bm x,y)\in\mathcal D_{\mathrm{cal}}}
\ell_{\mathrm{CE}}(f_{\bm{\theta}(\bm{\lambda})}(\bm x),y),
\end{aligned}
\label{eq:s_cal_ce}
\end{equation}
where $(\bm x,y)$ is an image--label pair, $f_{\bm{\theta}}$ is the classifier,
and $\ell_{\mathrm{CE}}$ is the cross-entropy loss.
The attention scores and client task vectors remain fixed during coefficient optimization.
Define the FedQCal coefficient family by
\begin{equation}
\mathcal F_+=\{\bm{\lambda}(\bm a):\bm a\ge\bm0\}.
\end{equation}
By the characterization in Sec.~\ref{app:implicit},
every feasible fixed-threshold FedQCal solution
$\bm{\lambda}^{\mathrm{fix}}$ belongs to $\mathcal F_+$.
Therefore,
\begin{equation}
\inf_{\bm{\lambda}\in\mathcal F_+}L_{\mathrm{cal}}(\bm{\lambda})
\le L_{\mathrm{cal}}(\bm{\lambda}^{\mathrm{fix}}),
\label{eq:s_ce_nesting}
\end{equation}
where $\inf$ denotes the greatest lower bound over the indicated set.
This comparison follows from membership in the coefficient family alone.
It imposes no ordering on the losses reached by finite optimization runs or on test accuracy.

On each region with a fixed clipping pattern, Eq.~\eqref{eq:s_coefficient_map} is affine.
Its Jacobian, the matrix of derivatives of $\bm{\lambda}$ with
respect to $\bm a$, has rank at most two.
Globally the family is piecewise affine and need not be a two-dimensional linear subspace:
clipping can produce a coefficient vector outside
$\operatorname{span}\{\hat{\bm\gamma},\tilde{\bm g}\}$.
Here $\operatorname{span}$ denotes the set of all linear combinations.
Nevertheless, clients with zero scores always receive zero weight,
and clients with identical score pairs always receive identical coefficients.
Calibration uses the full predictor to select within these restrictions.

\paragraph{Relative weights and merge scale.}
Before clipping becomes active, the coefficients equal $u\hat{\bm\gamma}+v\tilde{\bm g}$.
While clipping remains inactive, multiplying both coordinates by the same positive factor
rescales all coefficients and the complete parameter update,
without changing relative client weights.
Changing their ratio can also change relative weights when the
two score vectors are not proportional.
Normalized agreement gives equal scores to equally aligned clients,
whereas projection magnitude distinguishes the sizes of their aligned components.
If $\tilde{\bm g}=c\hat{\bm\gamma}$ for some $c\ge0$,
the unclipped coefficients depend only on $u+cv$:
the two coordinates then control a single scale.
Once some coefficients reach the cap, further scaling can also change relative weights,
because saturated coefficients remain fixed while others continue to grow.

\subsection{Calibration gradients and optimization}
\label{app:optimization}
Let $F(\bm a)=L_{\mathrm{cal}}(\bm{\lambda}(\bm a))$.
Where the merged-network loss is differentiable
and no coordinate lies at a clipping boundary, set $s_k=(\bm C^\top\bm a)_k$
and $\bm M_{\mathrm{clip}}(\bm a)=\operatorname{diag}(\mathbf1_{\{|s_k|<b\}})$.
Here $\operatorname{diag}$ forms a diagonal matrix from the listed entries,
and $\mathbf1_{\{E\}}$ equals one when condition $E$ holds and zero otherwise.
The chain rule gives
\begin{equation}
\nabla_{\bm a}F(\bm a)
=\bm C\bm M_{\mathrm{clip}}(\bm a)\nabla_{\bm{\lambda}}L_{\mathrm{cal}}(\bm{\lambda}(\bm a)).
\label{eq:s_chain_rule}
\end{equation}
Equivalently, with $r_k=\partial L_{\mathrm{cal}}/\partial\lambda_k$,
\begin{equation}
\nabla_{\bm a}F
=\sum_{k:\,|s_k|<b}
\begin{pmatrix}\hat\gamma_k\\\tilde g_k\end{pmatrix}r_k.
\label{eq:s_ce_gradient}
\end{equation}
The derivatives $r_k$ include all trainable parameters in $\Delta\bm{\theta}_k$,
including the classifier, through the actual parameter-merged network.
At a clipping boundary, the coefficient map is generally nondifferentiable,
and the implementation's automatic-differentiation convention determines the derivative used.

FedQCal forms a projected-gradient direction and searches along it:
\begin{equation}
\begin{aligned}
\bm d&=\Pi_{\mathbb{R}_+^2}\bigl(\bm a-\nabla F(\bm a)\bigr)-\bm a,\\
\bm a^+&=\bm a+\eta\bm d.
\end{aligned}
\end{equation}
Here $0<\eta\le1$ is the trial step size, $\bm a^+$ is the trial iterate,
and $\mathbb{R}_+^2=\{\bm a\in\mathbb{R}^2:\bm a\ge\bm0\}$ is the nonnegative quadrant.
The projection replaces negative coordinates by zero.
The direction $\bm d$ remains fixed during backtracking.
A trial is accepted only if its loss is finite, strictly decreases,
and satisfies the Armijo sufficient-decrease condition~\cite{armijo1966minimization}:
\begin{equation}
F(\bm a^+)\le F(\bm a)
+\sigma\nabla F(\bm a)^\top(\bm a^+-\bm a),
\quad 0<\sigma<1.
\label{eq:s_armijo}
\end{equation}
At each iteration, backtracking starts from $\eta=1$ and halves it,
with $\sigma=0.5$ and minimum step $2^{-20}$.
The run terminates if the projected direction vanishes,
the gradient or directional derivative is nonfinite,
a numerical stall occurs, or no acceptable trial is found at or above the minimum step.
Optimization permits at most 100 accepted updates and returns the last accepted iterate,
or the initialization if no update is accepted.
The CE objective is generally nonconvex;
sufficient decrease does not certify its global minimum or test-accuracy improvement.

\section{Reference Projection and Perturbations}
\label{app:projection}
\subsection{Projection identity}
If $h_k\le0$, then $\hat\gamma_k=\tilde g_k=0$
and Eq.~\eqref{eq:s_coefficient_map} gives $\lambda_k(\bm a)=0$.
For every remaining client $h_k=\tilde g_k$.
Consequently,
\begin{equation}
\left\langle \sum_k\lambda_k(\bm a)\Delta\bm W_k,\widehat{\bm R}\right\rangle_F
=\sum_k\lambda_k(\bm a)h_k
=\tilde{\bm g}^{\top}\bm{\lambda}(\bm a).
\label{eq:s_projection}
\end{equation}
This equals the induced second threshold and is nonnegative.
The support restriction is essential: the identity need not hold for an arbitrary
box-feasible coefficient vector, which could assign nonzero
weight to a negative-projection client.
A projection value does not control the components orthogonal to the reference
and is neither a cosine-similarity bound nor a distance bound.

\subsection{Proof of the target-projection bound}
\label{app:projection_proof}
Let arbitrary ideal matrices $\Delta\bm W_k^*,\Delta\bm W_{\mathrm{ref}}^*$ satisfy
$\|\Delta\bm W_k-\Delta\bm W_k^*\|_F\le\epsilon_k$
and $\|\Delta\bm W_{\mathrm{ref}}-\Delta\bm W_{\mathrm{ref}}^*\|_F\le\epsilon_{\mathrm{ref}}<\|\Delta\bm W_{\mathrm{ref}}^*\|_F$.
The normalization inequality is
\begin{align}
&\left\|\frac{\Delta\bm W_{\mathrm{ref}}}{\|\Delta\bm W_{\mathrm{ref}}\|_F}
-\frac{\Delta\bm W_{\mathrm{ref}}^*}{\|\Delta\bm W_{\mathrm{ref}}^*\|_F}\right\|_F\nonumber\\
&\le\frac{\|\Delta\bm W_{\mathrm{ref}}-\Delta\bm W_{\mathrm{ref}}^*\|_F}
{\|\Delta\bm W_{\mathrm{ref}}^*\|_F}\nonumber\\
&\quad+\frac{\bigl|\|\Delta\bm W_{\mathrm{ref}}^*\|_F-\|\Delta\bm W_{\mathrm{ref}}\|_F\bigr|}
{\|\Delta\bm W_{\mathrm{ref}}^*\|_F}\nonumber\\
&\le\frac{2\epsilon_{\mathrm{ref}}}{\|\Delta\bm W_{\mathrm{ref}}^*\|_F}.
\label{eq:s_normalize}
\end{align}
For $h_k^*=\left\langle \Delta\bm W_k^*,\Delta\bm W_{\mathrm{ref}}^*/\|\Delta\bm W_{\mathrm{ref}}^*\|_F\right\rangle_F$, this gives
\begin{align}
|h_k-h_k^*|
&\le\|\Delta\bm W_k-\Delta\bm W_k^*\|_F\nonumber\\
&\quad+\|\Delta\bm W_k^*\|_F
\left\|\widehat{\bm R}-\frac{\Delta\bm W_{\mathrm{ref}}^*}{\|\Delta\bm W_{\mathrm{ref}}^*\|_F}\right\|_F\nonumber\\
&\le\epsilon_k+2\|\Delta\bm W_k^*\|_F\epsilon_{\mathrm{ref}}/\|\Delta\bm W_{\mathrm{ref}}^*\|_F=:e_k.
\label{eq:s_project_error}
\end{align}
Define $\bm e=(e_1,\ldots,e_K)^\top$.
For any dual-induced $\bm{\lambda}\ge\bm0$, multiply the individual errors by $\lambda_k$
and use Eq.~\eqref{eq:s_projection} to obtain
\begin{align}
&\left\langle \sum_k\lambda_k\Delta\bm W_k^*,\frac{\Delta\bm W_{\mathrm{ref}}^*}{\|\Delta\bm W_{\mathrm{ref}}^*\|_F}\right\rangle_F\nonumber\\
&\ge\tilde{\bm g}^{\top}\bm{\lambda}-\sum_k\lambda_ke_k\nonumber\\
&\ge\tilde{\bm g}^{\top}\bm{\lambda}-\|\bm e\|_2\|\bm{\lambda}\|_2.
\end{align}
The error bounds are conditional quantities; the experiments do not estimate ideal target
summaries or certify numerical values of $\epsilon_k$ or $\epsilon_{\mathrm{ref}}$.

For any fixed coefficient vector and uncertainty radius $\varepsilon\ge0$,
let $\bm \xi\in\mathbb{R}^K$ denote an additive perturbation of the score vector.
The exact score-uncertainty identity is
\begin{equation}
\min_{\|\bm \xi\|_2\le\varepsilon}(\tilde{\bm g}+\bm \xi)^\top\bm{\lambda}
=\tilde{\bm g}^{\top}\bm{\lambda}-\varepsilon\|\bm{\lambda}\|_2.
\end{equation}
For nonzero $\bm{\lambda}$, equality follows by taking $\bm \xi=-\varepsilon\bm{\lambda}/\|\bm{\lambda}\|_2$.
The QP characterization minimizes the norm-based sensitivity
penalty for its selected constraints.
It does not imply that CE optimization maximizes this robust score,
nor does it establish stability of the learned coordinates
after replacing the calibration set.

\section{Parameter Merging and the Limits of the Scores}
\label{app:geometry}
Let $\bm W_K(\bm{\theta})$ and $\bm W_Q(\bm{\theta})$ denote the last attention block's key
and query weight matrices, and write $\bm W(\bm{\theta})=\bm W_K(\bm{\theta})^\top\bm W_Q(\bm{\theta})$
and $\bm W_0=\bm W(\bm{\theta}_0)$.
The indexed matrices satisfy $\bm W_{K,k}=\bm W_K(\bm{\theta}_k)$
and $\bm W_{Q,k}=\bm W_Q(\bm{\theta}_k)$, including $k=0$ for the shared initialization.
Write $\delta\bm K_k=\bm W_{K,k}-\bm W_{K,0}$ and $\delta\bm Q_k=\bm W_{Q,k}-\bm W_{Q,0}$.
Then the client summary is
\begin{equation}
\Delta\bm W_k=\bm W_{K,0}^{\top}\delta\bm Q_k
+\delta\bm K_k^{\top}\bm W_{Q,0}
+\delta\bm K_k^{\top}\delta\bm Q_k.
\end{equation}
Let $\bm M(\bm{\lambda})=\bm W(\bm{\theta}(\bm{\lambda}))-\bm W_0$ denote the actual change after parameter
merging, and $\bm Z(\bm{\lambda})=\sum_k\lambda_k\Delta\bm W_k$ the sum of client summaries.
Their discrepancy is exactly
\begin{align}
\bm D(\bm{\lambda})&=\bm M(\bm{\lambda})-\bm Z(\bm{\lambda})\nonumber\\
&=\sum_{i,j}(\lambda_i\lambda_j-\mathbf1_{\{i=j\}}\lambda_i)
\delta\bm K_i^{\top}\delta\bm Q_j.
\label{eq:s_discrepancy}
\end{align}
The discrepancy contains within-client and cross-client quadratic terms.
For dual-induced coefficients, Eq.~\eqref{eq:s_projection} implies
\begin{equation}
\left\langle \bm M(\bm{\lambda}),\widehat{\bm R}\right\rangle_F
\ge\tilde{\bm g}^{\top}\bm{\lambda}-\|\bm D(\bm{\lambda})\|_F.
\label{eq:s_merged_projection_bound}
\end{equation}
A coefficient-norm bound alone does not bound this discrepancy
without bounds on the client-update matrices.
We therefore measure it directly at the final calibrated models
for all 54 experimental conditions corresponding to Table~\ref{tab:fedqp_main}.
Table~\ref{tab:summary_discrepancy} reports the relative discrepancy
and the normalized lower bound.

\begin{table}[t]
\centering
\caption{Summary discrepancy at the final calibrated models
in the Table~\ref{tab:fedqp_main} settings ($n=5$).
Each dataset includes 18 conditions. All values are percentages.}
\label{tab:summary_discrepancy}
\small
\setlength{\tabcolsep}{5pt}
\begin{tabular}{lcc}
\toprule
Dataset
& $\displaystyle \frac{\|D\|_F}{\|M\|_F}$
& $\displaystyle
\frac{\tilde{\bm g}^{\top}\lambda-\|D(\lambda)\|_F}
     {\tilde{\bm g}^{\top}\lambda}$ \\
& Median (max.) & Min. \\
\midrule
CIFAR-10  & 2.23 (11.08)  & 87.08 \\
CIFAR-100 & 11.00 (17.84) & 78.17 \\
SVHN      & 4.25 (9.16)   & 86.98 \\
\bottomrule
\end{tabular}
\end{table}

The lower bound remains positive in every condition and retains
at least 78.17\% of the weighted-summary reference projection.
Thus, although the two summaries differ, the discrepancy does
not eliminate positive reference alignment in the evaluated settings.

Moreover, attention summaries do not identify classifier behavior.
Models with the same key and query matrices can have different
classifier heads and different predictions.
For example, permuting classifier output rows leaves all attention summaries unchanged
while changing label assignments.
This observation concerns the information contained in the scores;
it is not an assertion that such a permutation occurs during the experiments.
FedQCal evaluates the complete merged predictor to select its two coordinates,
but the scores still determine which coefficient vectors those coordinates can represent.

\section{Implementation Details}
\label{sec:implementation}

The following details apply to the ViT-Tiny experiments.
Settings specific to additional architectures are described
alongside their results.

\paragraph{Models and data.}
We use the timm checkpoint
\texttt{vit\_}\allowbreak\texttt{tiny\_}\allowbreak\texttt{patch16\_}\allowbreak\texttt{224.augreg\_in21k\_ft\_in1k}.
Client and reference training use random resized cropping
to $224\times224$ and random horizontal flipping
with torchvision's default settings;
horizontal flipping is omitted for SVHN.
Calibration and evaluation instead use
\texttt{Resize(224)} followed by \texttt{ToTensor}.
All stages normalize each channel with mean and standard
deviation both equal to $0.5$.

For each training seed, five images per class are sampled
without replacement before client partitioning, using NumPy's
\texttt{default\_rng} with seed $\texttt{seed}+100$.
For each class, the remaining indices are shuffled,
client proportions are drawn from a symmetric
$\operatorname{Dirichlet}(\alpha,\ldots,\alpha)$ distribution,
and integer client counts are sampled from the corresponding
multinomial distribution.
The partition generator starts with seed $\texttt{seed}+200$.
If any client receives no samples, the entire partition is
resampled with the seed incremented by one, for at most
1,000 attempts.

The one-image-per-class pool is sampled without replacement
from the same five-image holdout using seed $\texttt{seed}+500$;
unused held-out images remain excluded from client training.
Both reference training and coefficient calibration use the
selected pool.
The fixed reference-training sample list described in the main
paper is generated using seed $\texttt{seed}+300$.
Thus, changing the server-data budget affects both reference
training and coefficient calibration. 
Table~\ref{tab:supp_calibration} summarizes the total number
of distinct server images for each per-class budget.

\begin{table}[t]
\centering
\caption{\textbf{Labeled server-data budgets.}
Entries give the total number of distinct server images.}
\label{tab:supp_calibration}
\small
\begin{tabular}{lrr}
\toprule
Dataset & One per class & Five per class\\
\midrule
CIFAR-10 & 10 & 50\\
CIFAR-100 & 100 & 500\\
SVHN & 10 & 50\\
\bottomrule
\end{tabular}
\end{table}

\begin{table*}[t]
\centering
\caption{\textbf{One-shot federated learning with ViT-S.}
Test accuracy (\%), mean $\pm$ SD over three seeds.
Server-supervised methods use five labeled server images per class.
Bold and underlined values indicate the highest and second-highest
displayed mean accuracies within each client-count block and column;
ties share a rank.}
\label{tab:supp_vits_n5}
\small
\setlength{\tabcolsep}{4pt}
\resizebox{\textwidth}{!}{%
\begin{tabular}{lccccccccc}
\toprule
& \multicolumn{3}{c}{CIFAR-10}
& \multicolumn{3}{c}{CIFAR-100}
& \multicolumn{3}{c}{SVHN}\\
\cmidrule(lr){2-4}\cmidrule(lr){5-7}\cmidrule(lr){8-10}
Method
& $\alpha=0.05$ & $\alpha=0.10$ & $\alpha=0.50$
& $\alpha=0.05$ & $\alpha=0.10$ & $\alpha=0.50$
& $\alpha=0.05$ & $\alpha=0.10$ & $\alpha=0.50$\\
\midrule

\multicolumn{10}{l}{\emph{10 clients}}\\

Uniform averaging
& $69.65\pm1.51$ & $82.33\pm9.00$ & $96.26\pm0.19$
& $44.76\pm1.60$ & $57.36\pm3.10$ & $79.27\pm0.51$
& $23.34\pm5.41$ & $27.18\pm0.87$ & $53.78\pm1.21$\\

FedAvg
& $79.80\pm3.43$ & $84.58\pm6.63$ & $\underline{96.33\pm0.29}$
& $45.88\pm1.88$ & $57.76\pm2.06$ & $79.35\pm0.54$
& $23.35\pm3.19$ & $31.56\pm5.68$ & $54.30\pm1.37$\\

Ensemble
& $70.68\pm4.68$ & $82.37\pm9.33$ & $96.28\pm0.16$
& $45.58\pm1.40$ & $58.17\pm3.01$ & $79.55\pm0.58$
& $24.74\pm4.49$ & $29.84\pm3.04$ & $\underline{56.00\pm1.15}$\\

TIES-Merging
& $68.15\pm2.22$ & $82.34\pm8.54$ & $96.31\pm0.17$
& $45.29\pm0.84$ & $58.26\pm3.62$ & $79.64\pm0.73$
& $24.26\pm5.48$ & $27.19\pm6.71$ & $53.34\pm1.64$\\

\midrule

Reference-only
& $84.01\pm2.17$ & $84.01\pm2.17$ & $84.01\pm2.17$
& $\underline{67.73\pm0.95}$ & $\underline{67.73\pm0.95}$ & $67.73\pm0.95$
& $15.87\pm1.08$ & $15.87\pm1.08$ & $15.87\pm1.08$\\

FedAvg + calibration
& $\mathbf{94.38\pm0.20}$ & $\mathbf{94.66\pm0.96}$ & $95.78\pm0.21$
& $\mathbf{77.69\pm0.59}$ & $\mathbf{78.77\pm0.60}$ & $\mathbf{81.45\pm0.49}$
& $\underline{26.61\pm2.26}$ & $\underline{31.70\pm1.50}$ & $42.54\pm2.31$\\

FedQCal
& $\underline{86.64\pm3.28}$ & $\underline{92.03\pm5.34}$ & $\mathbf{96.46\pm0.21}$
& $59.32\pm3.47$ & $65.18\pm3.52$ & $\underline{81.21\pm0.36}$
& $\mathbf{43.41\pm4.61}$ & $\mathbf{45.06\pm3.69}$ & $\mathbf{58.39\pm2.36}$\\

\midrule\midrule
\multicolumn{10}{l}{\emph{100 clients}}\\

Uniform averaging
& $60.46\pm5.52$ & $88.32\pm3.84$ & $95.52\pm0.21$
& $26.19\pm1.38$ & $47.30\pm1.78$ & $73.75\pm0.42$
& $16.60\pm0.42$ & $31.19\pm2.03$ & $51.55\pm0.96$\\

FedAvg
& $86.90\pm2.48$ & $91.96\pm1.49$ & $95.63\pm0.16$
& $28.48\pm1.11$ & $49.04\pm1.63$ & $74.07\pm0.33$
& $\underline{25.32\pm1.83}$ & $37.95\pm5.42$ & $53.12\pm0.43$\\

Ensemble
& $63.80\pm4.60$ & $89.45\pm3.77$ & $\underline{95.71\pm0.22}$
& $28.12\pm1.57$ & $49.76\pm1.88$ & $74.78\pm0.44$
& $18.84\pm0.55$ & $\underline{39.98\pm1.19}$ & $\mathbf{55.82\pm1.08}$\\

TIES-Merging
& $56.91\pm4.32$ & $87.58\pm5.07$ & $95.69\pm0.21$
& $29.72\pm1.57$ & $51.78\pm1.79$ & $75.77\pm0.21$
& $16.01\pm2.15$ & $31.21\pm4.17$ & $50.07\pm1.10$\\

\midrule

Reference-only
& $84.61\pm1.30$ & $84.61\pm1.30$ & $84.61\pm1.30$
& $67.68\pm0.70$ & $67.68\pm0.70$ & $67.68\pm0.70$
& $16.04\pm0.82$ & $16.04\pm0.82$ & $16.04\pm0.82$\\

FedAvg + calibration
& $\mathbf{93.48\pm0.41}$ & $\underline{93.87\pm0.28}$ & $95.03\pm0.28$
& $\mathbf{75.22\pm0.68}$ & $\mathbf{76.28\pm0.78}$ & $\underline{78.31\pm0.74}$
& $24.15\pm1.65$ & $27.71\pm1.56$ & $37.40\pm2.63$\\

FedQCal
& $\underline{90.58\pm1.98}$ & $\mathbf{94.74\pm0.58}$ & $\mathbf{95.82\pm0.29}$
& $\underline{69.97\pm1.84}$ & $\underline{75.57\pm0.67}$ & $\mathbf{78.96\pm0.41}$
& $\mathbf{44.63\pm2.71}$ & $\mathbf{49.66\pm0.61}$ & $\underline{53.74\pm0.42}$\\

\bottomrule
\end{tabular}%
}
\end{table*}

\paragraph{Calibration and controls.}
During standard FedQCal calibration, the client task vectors,
reference model, and alignment scores remain fixed.
All trainable parameters of each client, including its classifier,
share that client's merge coefficient.

The independent-head variant initializes the classifier weights
and biases from the same initial merged model as standard FedQCal.
The Transformer block retains the original coefficient
parameterization.
The two coordinates use the projected gradient direction
described in Appendix~\ref{app:optimization}, while the classifier weights and biases
use their unscaled negative gradients.
A single common step size updates both parts and is accepted
using their joint calibration CE and the Armijo settings
in Appendix~\ref{app:optimization}.
No momentum, weight decay, or additional regularization is
used during this joint calibration.

For the random-direction control, we draw
\begin{align}
&\bm z_1,\bm z_2
\overset{\mathrm{iid}}{\sim}\mathcal N(\bm 0,I_K),
\notag\\
&\bm r_1
=\|\hat{\bm\gamma}\|_2
 \frac{|\bm z_1|}{\|\bm z_1\|_2},
\qquad
\bm r_2
=\|\tilde{\bm g}\|_2
 \frac{|\bm z_2|}{\|\bm z_2\|_2}.
\label{eq:random_directions}
\end{align}
Here absolute values are taken elementwise.
These are independent uniform directions on the nonnegative
unit sphere, scaled to match the two score-vector norms.
The two coordinates are initialized as
\begin{align}
\bm a_0
=
\left(
\frac{1}{2\sum_k r_{1,k}},
\frac{1}{2\sum_k r_{2,k}}
\right)^\top. \notag
\end{align}
Here $r_{j,k}$ denotes the $k$th component of $\bm r_j$,
for $j=1,2$ and $k=1,\ldots,K$.
This matches the initial coefficient sum of FedQCal but generally
produces a different initial merged model.
The coefficient cap, nonnegative coordinate domain, calibration
examples, and Armijo settings are unchanged.

For each pair index $p\in\{0,\ldots,4\}$, we use NumPy's
default\_rng generator, initialized through SeedSequence
with the entropy tuple
$(20261106,\,503,\,K,\,\mathrm{seed},\,p,\,1)$,
where $\mathrm{seed}\in\{42,43,44\}$ is the training seed.
The base Gaussian draws are shared across datasets,
$\alpha$, and server-data budgets at fixed $K$,
training seed, and pair index.
All five pairs are retained without selection.
Their test accuracies are averaged within each training seed;
the reported mean and sample standard deviation are then
computed over the three seed-level averages.

\paragraph{Degenerate cases.}
If the reference summary $\Delta\bm W_{\mathrm{ref}}$ is zero,
or if both score vectors vanish, the server returns the
sample-weighted FedAvg model before calibration, using
\begin{align}
\lambda_k=\frac{N_k}{\sum_jN_j}, \notag
\end{align}
where $N_k$ is the number of training examples at client $k$.
This fallback is separate from the QP family analyzed above.

\begin{table*}[t]
\centering
\caption{\textbf{Comparison with one labeled server image per class.}
Test accuracy (\%), mean $\pm$ SD over three seeds.}
\label{tab:supp_n1}
\small
\setlength{\tabcolsep}{4pt}
\resizebox{\textwidth}{!}{%
\begin{tabular}{lccccccccc}
\toprule
& \multicolumn{3}{c}{CIFAR-10}
& \multicolumn{3}{c}{CIFAR-100}
& \multicolumn{3}{c}{SVHN}\\
\cmidrule(lr){2-4}\cmidrule(lr){5-7}\cmidrule(lr){8-10}
Method
& $\alpha=0.05$ & $\alpha=0.10$ & $\alpha=0.50$
& $\alpha=0.05$ & $\alpha=0.10$ & $\alpha=0.50$
& $\alpha=0.05$ & $\alpha=0.10$ & $\alpha=0.50$\\
\midrule

\multicolumn{10}{l}{\emph{ViT-Tiny, 10 clients}}\\

FedAvg
& $71.46\pm3.60$ & $76.21\pm5.27$ & $91.92\pm0.39$
& $37.84\pm2.13$ & $48.45\pm1.20$ & $69.95\pm0.60$
& $21.67\pm2.67$ & $23.72\pm4.37$ & $46.78\pm1.30$\\

Reference-only
& $49.61\pm6.82$ & $49.61\pm6.82$ & $49.61\pm6.82$
& $26.50\pm1.12$ & $26.50\pm1.12$ & $26.50\pm1.12$
& $11.34\pm0.95$ & $11.34\pm0.95$ & $11.34\pm0.95$\\

FedAvg + calibration
& $81.59\pm4.34$ & $84.35\pm1.16$ & $90.80\pm0.80$
& $48.54\pm1.37$ & $54.90\pm0.44$ & $67.13\pm0.72$
& $20.81\pm3.81$ & $25.54\pm3.94$ & $35.75\pm4.48$\\

FedQCal
& $74.93\pm5.08$ & $85.58\pm3.01$ & $92.05\pm0.27$
& $51.31\pm3.53$ & $58.10\pm4.04$ & $71.63\pm0.74$
& $39.01\pm5.48$ & $43.46\pm3.32$ & $52.92\pm0.24$\\

\midrule
\multicolumn{10}{l}{\emph{ViT-Tiny, 100 clients}}\\

FedAvg
& $76.96\pm3.34$ & $84.46\pm2.42$ & $91.03\pm0.30$
& $22.56\pm1.94$ & $38.34\pm0.35$ & $62.95\pm0.29$
& $18.57\pm4.59$ & $30.03\pm3.86$ & $46.72\pm0.36$\\

Reference-only
& $47.43\pm6.00$ & $47.43\pm6.00$ & $47.43\pm6.00$
& $26.22\pm1.09$ & $26.22\pm1.09$ & $26.22\pm1.09$
& $11.25\pm1.04$ & $11.25\pm1.04$ & $11.25\pm1.04$\\

FedAvg + calibration
& $80.66\pm4.47$ & $83.65\pm3.53$ & $89.43\pm0.75$
& $40.65\pm1.26$ & $45.48\pm1.41$ & $56.95\pm0.58$
& $19.55\pm3.16$ & $23.95\pm3.26$ & $36.48\pm2.20$\\

FedQCal
& $83.06\pm8.36$ & $86.97\pm3.11$ & $90.52\pm0.96$
& $50.87\pm1.53$ & $57.16\pm0.53$ & $66.20\pm0.30$
& $39.12\pm2.34$ & $44.46\pm0.77$ & $48.33\pm1.69$\\

\midrule
\multicolumn{10}{l}{\emph{ViT-S, 10 clients}}\\

FedAvg
& $79.80\pm3.43$ & $84.58\pm6.63$ & $96.33\pm0.29$
& $45.88\pm1.88$ & $57.76\pm2.06$ & $79.35\pm0.54$
& $23.35\pm3.19$ & $31.56\pm5.68$ & $54.30\pm1.37$\\

Reference-only
& $62.20\pm5.39$ & $62.20\pm5.39$ & $62.20\pm5.39$
& $37.41\pm3.00$ & $37.41\pm3.00$ & $37.41\pm3.00$
& $10.87\pm0.83$ & $10.87\pm0.83$ & $10.87\pm0.83$\\

FedAvg + calibration
& $92.23\pm1.22$ & $93.70\pm0.96$ & $96.36\pm0.23$
& $64.67\pm1.05$ & $69.58\pm0.69$ & $78.79\pm1.17$
& $22.64\pm3.43$ & $29.42\pm2.48$ & $43.62\pm2.81$\\

FedQCal
& $85.78\pm3.78$ & $91.83\pm6.38$ & $96.57\pm0.25$
& $58.90\pm3.03$ & $65.31\pm4.01$ & $80.90\pm0.72$
& $43.08\pm5.17$ & $45.32\pm3.54$ & $57.72\pm3.24$\\

\midrule
\multicolumn{10}{l}{\emph{ViT-S, 100 clients}}\\

FedAvg
& $86.90\pm2.48$ & $91.96\pm1.49$ & $95.63\pm0.16$
& $28.48\pm1.11$ & $49.04\pm1.63$ & $74.07\pm0.33$
& $25.32\pm1.83$ & $37.95\pm5.42$ & $53.12\pm0.43$\\

Reference-only
& $62.77\pm5.56$ & $62.77\pm5.56$ & $62.77\pm5.56$
& $36.95\pm2.21$ & $36.95\pm2.21$ & $36.95\pm2.21$
& $11.16\pm1.14$ & $11.16\pm1.14$ & $11.16\pm1.14$\\

FedAvg + calibration
& $91.92\pm0.63$ & $92.95\pm0.43$ & $95.40\pm0.12$
& $55.45\pm1.84$ & $60.64\pm1.87$ & $70.99\pm1.21$
& $21.03\pm1.10$ & $26.39\pm1.59$ & $40.74\pm0.48$\\

FedQCal
& $89.96\pm2.23$ & $94.46\pm0.45$ & $95.35\pm0.84$
& $69.88\pm1.82$ & $75.49\pm0.76$ & $78.87\pm0.52$
& $44.48\pm0.67$ & $47.81\pm3.95$ & $53.31\pm0.42$\\

\bottomrule
\end{tabular}%
}
\end{table*}

\begin{table*}[t]
\centering
\caption{\textbf{Extended ablations.}
Test accuracy (\%), mean $\pm$ SD over three seeds.
$n$ denotes labeled server images per class.}
\label{tab:supp_ablations}
\small
\setlength{\tabcolsep}{4pt}
\resizebox{\textwidth}{!}{%
\begin{tabular}{lccccccccc}
\toprule
& \multicolumn{3}{c}{CIFAR-10}
& \multicolumn{3}{c}{CIFAR-100}
& \multicolumn{3}{c}{SVHN}\\
\cmidrule(lr){2-4}\cmidrule(lr){5-7}\cmidrule(lr){8-10}
Setting
& $\alpha=0.05$ & $\alpha=0.10$ & $\alpha=0.50$
& $\alpha=0.05$ & $\alpha=0.10$ & $\alpha=0.50$
& $\alpha=0.05$ & $\alpha=0.10$ & $\alpha=0.50$\\
\midrule

\multicolumn{10}{l}{\emph{ViT-Tiny, 100 clients, $n=5$}}\\

FedQCal
& $89.49\pm0.62$ & $88.49\pm0.13$ & $90.51\pm1.59$
& $51.12\pm1.61$ & $57.30\pm0.68$ & $66.26\pm0.08$
& $37.91\pm3.44$ & $43.72\pm1.95$ & $47.16\pm1.97$\\

Initialization only
& $67.56\pm4.97$ & $83.21\pm1.50$ & $90.88\pm0.30$
& $21.52\pm1.70$ & $37.15\pm0.46$ & $62.71\pm0.29$
& $19.51\pm3.48$ & $28.97\pm0.09$ & $45.53\pm0.55$\\

Random directions
& $84.51\pm2.36$ & $87.91\pm0.64$ & $90.21\pm1.63$
& $49.37\pm1.71$ & $56.21\pm0.63$ & $65.89\pm0.29$
& $31.64\pm7.01$ & $41.44\pm3.18$ & $46.76\pm2.13$\\

Independent head
& $85.07\pm0.38$ & $86.93\pm0.60$ & $89.78\pm0.29$
& $58.76\pm0.83$ & $60.49\pm1.14$ & $65.25\pm0.72$
& $25.85\pm1.78$ & $28.55\pm1.64$ & $36.38\pm1.63$\\

FedAvg + calibration
& $85.38\pm0.34$ & $86.39\pm0.42$ & $89.11\pm0.24$
& $59.11\pm1.13$ & $60.55\pm1.25$ & $63.83\pm0.99$
& $23.85\pm1.10$ & $26.41\pm1.20$ & $33.18\pm2.04$\\

\midrule
\multicolumn{10}{l}{\emph{ViT-Tiny, 100 clients, $n=1$}}\\

FedQCal
& $83.06\pm8.36$ & $86.97\pm3.11$ & $90.52\pm0.96$
& $50.87\pm1.53$ & $57.16\pm0.53$ & $66.20\pm0.30$
& $39.12\pm2.34$ & $44.46\pm0.77$ & $48.33\pm1.69$\\

Initialization only
& $67.78\pm5.18$ & $83.42\pm1.58$ & $90.92\pm0.24$
& $21.46\pm1.60$ & $37.08\pm0.49$ & $62.70\pm0.30$
& $19.43\pm3.98$ & $28.76\pm0.19$ & $45.71\pm0.59$\\

Random directions
& $75.83\pm15.30$ & $86.65\pm2.36$ & $90.35\pm0.71$
& $49.34\pm1.55$ & $56.24\pm0.58$ & $65.85\pm0.27$
& $37.45\pm1.24$ & $42.88\pm0.18$ & $48.02\pm1.63$\\

Independent head
& $79.82\pm5.57$ & $84.85\pm2.76$ & $90.05\pm0.45$
& $40.40\pm0.98$ & $46.31\pm1.27$ & $59.98\pm0.58$
& $22.38\pm1.16$ & $27.81\pm0.28$ & $40.37\pm1.00$\\

FedAvg + calibration
& $80.66\pm4.47$ & $83.65\pm3.53$ & $89.43\pm0.75$
& $40.65\pm1.26$ & $45.48\pm1.41$ & $56.95\pm0.58$
& $19.55\pm3.16$ & $23.95\pm3.26$ & $36.48\pm2.20$\\

\midrule
\multicolumn{10}{l}{\emph{ViT-S, 10 clients, $n=5$}}\\

FedQCal
& $86.64\pm3.28$ & $92.03\pm5.34$ & $96.46\pm0.21$
& $59.32\pm3.47$ & $65.18\pm3.52$ & $81.21\pm0.36$
& $43.41\pm4.61$ & $45.06\pm3.69$ & $58.39\pm2.36$\\

Initialization only
& $79.61\pm4.63$ & $89.26\pm6.80$ & $96.30\pm0.21$
& $45.29\pm1.77$ & $57.72\pm2.90$ & $79.38\pm0.56$
& $25.08\pm3.98$ & $30.02\pm2.87$ & $54.36\pm0.99$\\

Random directions
& $74.14\pm4.28$ & $86.25\pm6.09$ & $96.31\pm0.35$
& $50.76\pm2.78$ & $60.22\pm4.06$ & $79.74\pm0.36$
& $38.56\pm8.08$ & $40.11\pm4.60$ & $56.59\pm2.28$\\

Independent head
& $94.71\pm0.27$ & $94.93\pm1.06$ & $96.33\pm0.30$
& $78.20\pm0.58$ & $79.29\pm0.44$ & $82.19\pm0.23$
& $28.04\pm1.83$ & $34.05\pm2.58$ & $44.76\pm0.95$\\

FedAvg + calibration
& $94.38\pm0.20$ & $94.66\pm0.96$ & $95.78\pm0.21$
& $77.69\pm0.59$ & $78.77\pm0.60$ & $81.45\pm0.49$
& $26.61\pm2.26$ & $31.70\pm1.50$ & $42.54\pm2.31$\\

\midrule
\multicolumn{10}{l}{\emph{ViT-S, 10 clients, $n=1$}}\\

FedQCal
& $85.78\pm3.78$ & $91.83\pm6.38$ & $96.57\pm0.25$
& $58.90\pm3.03$ & $65.31\pm4.01$ & $80.90\pm0.72$
& $43.08\pm5.17$ & $45.32\pm3.54$ & $57.72\pm3.24$\\

Initialization only
& $79.53\pm4.49$ & $89.14\pm6.73$ & $96.31\pm0.20$
& $45.37\pm1.79$ & $57.77\pm2.98$ & $79.38\pm0.57$
& $24.69\pm4.10$ & $29.60\pm3.10$ & $54.26\pm1.01$\\

Random directions
& $71.56\pm5.87$ & $85.36\pm8.07$ & $96.38\pm0.16$
& $50.55\pm3.23$ & $60.10\pm4.05$ & $79.43\pm0.45$
& $38.56\pm5.61$ & $39.18\pm3.71$ & $55.53\pm2.77$\\

Independent head
& $93.48\pm0.48$ & $94.74\pm1.17$ & $96.71\pm0.09$
& $67.27\pm1.16$ & $71.91\pm0.82$ & $79.98\pm1.01$
& $23.22\pm1.95$ & $30.61\pm0.20$ & $46.88\pm0.98$\\

FedAvg + calibration
& $92.23\pm1.22$ & $93.70\pm0.96$ & $96.36\pm0.23$
& $64.67\pm1.05$ & $69.58\pm0.69$ & $78.79\pm1.17$
& $22.64\pm3.43$ & $29.42\pm2.48$ & $43.62\pm2.81$\\

\midrule
\multicolumn{10}{l}{\emph{ViT-S, 100 clients, $n=5$}}\\

FedQCal
& $90.58\pm1.98$ & $94.74\pm0.58$ & $95.82\pm0.29$
& $69.97\pm1.84$ & $75.57\pm0.67$ & $78.96\pm0.41$
& $44.63\pm2.71$ & $49.66\pm0.61$ & $53.74\pm0.42$\\

Initialization only
& $77.32\pm3.68$ & $90.60\pm2.39$ & $95.57\pm0.20$
& $27.22\pm1.22$ & $48.07\pm1.65$ & $73.91\pm0.43$
& $22.35\pm2.22$ & $36.10\pm3.90$ & $52.22\pm0.79$\\

Random directions
& $79.53\pm9.11$ & $93.64\pm0.69$ & $95.73\pm0.27$
& $67.58\pm1.26$ & $73.80\pm0.77$ & $78.56\pm0.47$
& $39.44\pm2.80$ & $46.45\pm0.42$ & $52.91\pm0.60$\\

Independent head
& $93.84\pm0.36$ & $94.94\pm0.23$ & $95.90\pm0.28$
& $73.91\pm0.78$ & $75.10\pm0.70$ & $77.90\pm0.48$
& $27.67\pm1.08$ & $31.61\pm1.30$ & $42.47\pm1.44$\\

FedAvg + calibration
& $93.48\pm0.41$ & $93.87\pm0.28$ & $95.03\pm0.28$
& $75.22\pm0.68$ & $76.28\pm0.78$ & $78.31\pm0.74$
& $24.15\pm1.65$ & $27.71\pm1.56$ & $37.40\pm2.63$\\

\midrule
\multicolumn{10}{l}{\emph{ViT-S, 100 clients, $n=1$}}\\

FedQCal
& $89.96\pm2.23$ & $94.46\pm0.45$ & $95.35\pm0.84$
& $69.88\pm1.82$ & $75.49\pm0.76$ & $78.87\pm0.52$
& $44.48\pm0.67$ & $47.81\pm3.95$ & $53.31\pm0.42$\\

Initialization only
& $77.32\pm3.77$ & $90.59\pm2.57$ & $95.56\pm0.18$
& $27.12\pm1.27$ & $47.99\pm1.59$ & $73.90\pm0.45$
& $22.74\pm2.35$ & $36.02\pm4.32$ & $52.37\pm0.76$\\

Random directions
& $86.75\pm2.18$ & $94.17\pm0.28$ & $95.30\pm0.79$
& $67.58\pm1.26$ & $73.69\pm0.76$ & $78.47\pm0.54$
& $36.36\pm4.52$ & $45.31\pm3.13$ & $52.34\pm0.59$\\

Independent head
& $93.06\pm0.47$ & $95.20\pm0.04$ & $96.21\pm0.12$
& $57.51\pm1.56$ & $62.85\pm2.32$ & $74.08\pm1.40$
& $24.63\pm0.63$ & $31.54\pm1.10$ & $46.37\pm0.07$\\

FedAvg + calibration
& $91.92\pm0.63$ & $92.95\pm0.43$ & $95.40\pm0.12$
& $55.45\pm1.84$ & $60.64\pm1.87$ & $70.99\pm1.21$
& $21.03\pm1.10$ & $26.39\pm1.59$ & $40.74\pm0.48$\\

\bottomrule
\end{tabular}%
}
\end{table*}

\section{Additional Experimental Results}
\label{sec:additional_results}
We extend the main comparisons in Table~\ref{tab:fedqp_main} and the component
ablations in Table~\ref{tab:fedqp_ablations} to additional backbone, client-count,
and server-data settings.

\subsection{Additional comparisons}
\label{sec:additional_comparisons}

Table~\ref{tab:supp_vits_n5} extends Table~\ref{tab:fedqp_main}
to ViT-S with five labeled server images per class.
We use the timm checkpoint
\texttt{vit\_\allowbreak{}small\_\allowbreak{}%
patch16\_\allowbreak{}224.\allowbreak{}%
augreg\_\allowbreak{}in21k\_\allowbreak{}%
ft\_\allowbreak{}in1k}, retaining the training and
calibration settings in Sec.~\ref{sec:protocol}
and Sec.~\ref{sec:implementation}.
FedQCal's gains over Uniform averaging, FedAvg, and
TIES-Merging extend to ViT-S.
It also exceeds Ensemble except on SVHN with 100 clients
at $\alpha=0.50$.
Its advantage over FedAvg + calibration is particularly
clear on SVHN, with gains of 13.36--21.95 percentage points,
although Reference-only achieves only about 16\% accuracy.
These results support using reference adaptation to guide
the combination of client knowledge even when the reference
is a poor standalone classifier.

Table~\ref{tab:supp_n1} extends the main comparison to one
labeled server image per class for both backbones and client
counts; overlapping ViT-Tiny results from Table~\ref{tab:fedqp_ablations} are repeated
to facilitate comparison.
FedQCal achieves higher mean accuracy than Reference-only
throughout the table and improves on FedAvg except on
CIFAR-10 with 100 clients at $\alpha=0.50$.
Its gains over FedAvg + calibration reach
7.89--14.85 percentage points on CIFAR-100 with 100 clients
and 11.85--23.46 points on SVHN.
These results support using scarce server labels to select
aggregation coefficients while retaining client knowledge.
At $\alpha=0.05$ and $0.10$, increasing the client count
from 10 to 100 reduces FedQCal's accuracy by at most
0.93 percentage points, with improvements in some settings.
Its accuracy change is more favorable than that of
FedAvg + calibration in every matched comparison at these
two skew levels.
Under our client-count-specific training schedules,
these results support applying two-coordinate calibration
to a larger heterogeneous client population even with
one labeled server image per class.

\subsection{Additional ablations}
\label{sec:additional_ablations}

Table~\ref{tab:supp_ablations} extends the coefficient-calibration,
reference-guided weighting, and independent-classifier
adaptation ablations in Table~\ref{tab:fedqp_ablations} to
ViT-Tiny with 100 clients and ViT-S with 10 and 100 clients,
using both server-data budgets.
The variants follow the protocols in Sec.~\ref{sec:implementation}.

\paragraph{Coefficient calibration.}
The initialization-only control retains the same reference,
scores, and coefficient map, testing whether prediction-based
selection improves on normalized-score initialization.
At $\alpha=0.05$ and $0.10$, calibration improves mean
accuracy throughout the evaluated settings, with larger
gains than at $\alpha=0.50$ when the other conditions
are held fixed.
These results support CE-based coefficient calibration from
normalized-score initialization, particularly under stronger
label skew.

\paragraph{Reference-guided weighting.}
The norm-matched random control preserves the two-coordinate
parameterization and optimization protocol, testing whether
these features alone reproduce reference-guided performance.
FedQCal exceeds the random-direction control in mean accuracy
throughout Table~\ref{tab:supp_ablations}, with larger gains at
$\alpha=0.05$ and $0.10$ than at $\alpha=0.50$ within each
matched comparison.
This supports using task information from the reference to
structure coefficient selection beyond low dimensionality
alone, including when the coefficient vector contains
100 client weights.
Because the random control changes both initialization and
search directions, the evidence concerns their combined
reference-guided construction.

\paragraph{Independent classifier adaptation.}
The independent-head control starts from the same initial
merged model and jointly optimizes the two coordinates and
all classifier weights and biases, while retaining the
original Transformer-block parameterization.
With 100 clients and one labeled server image per class,
FedQCal achieves higher mean accuracy than this control
on CIFAR-100 and SVHN for both backbones and all label-skew levels.
These results support retaining shared coefficient control
of the classifier when calibrating many client models
from scarce server labels in these datasets.

\paragraph{Server-data budget.}
We compare $n=5$ and $n=1$ across all 27 matched settings
in Table~\ref{tab:supp_ablations}, weighting each setting equally.
The mean absolute change in test accuracy is 0.70 percentage
points for FedQCal, compared with 4.71 for independent
classifier adaptation and 5.46 for FedAvg + calibration.
These results indicate lower average sensitivity to the
server-data budget, supporting coefficient calibration
with scarce server supervision.

\end{document}